\documentclass[final,5p,times,twocolumn]{elsarticle}

\usepackage{amssymb}
\usepackage{amsmath}
\usepackage[colorlinks]{hyperref}
\usepackage{amsmath}
\usepackage{booktabs}
\usepackage{amsfonts}
\usepackage{graphicx}
\usepackage{pifont}
\usepackage{amssymb}
\usepackage{xcolor}
\usepackage{subcaption}
\usepackage{gensymb}
\usepackage{amsmath}
\DeclareMathOperator*{\argmax}{arg\,max}

\usepackage{comment}
\usepackage[justification=centering]{caption}
\usepackage{xurl}     % improves URL line-breaking (optional but helpful)
\journal{Robotics and Autonomous Systems}

\begin{document}

\begin{frontmatter}


%% use the ead command for the email address,74

\title{Beyond Coverage Path Planning: Can UAV Swarms Perfect Scattered Regions Inspections?}

\author{Socratis Gkelios$^{a,b}$, Savvas D. Apostolidis$^{a,b}$, Pavlos Ch. Kapoutsis$^{b}$, Elias B. Kosmatopoulos$^{a,b}$  and Athanasios Ch. Kapoutsis$^{a, b}$}
%  \thanks{Manuscript received: - ; Revised -; Accepted- .}
%   %Use only for final RAL version 
%  \thanks{This project has received funding from the European Commission under the European
% Union’s Horizon 2020 research and innovation programme under grant agreement no 101073952 (PERIVALLON). We also gratefully acknowledge the support of NVIDIA Corporation with the donation of GPUs used for this research. (Corresponding author: Socratis Gkelios)}
%  %Use only for final RAL version
% 	\thanks{$^{1}$ Socratis Gkelios, Savvas D. Apostolidis and Elias B. Kosmatopoulos are with Department of Electrical and Computer Engineering,    Democritus University of Thrace, Xanthi, Greece and Information Technologies Institute, The Centre for Research \& Technology, Hellas, Thessaloniki, Greece (sgkelios@iti.gr, sapostol@iti.gr, kosmatop@iti.gr)}%
% 	\thanks{$^{2}$ Pavlos Ch. Kapoutsis and Athanasios Ch. Kapoutsis are with Information Technologies Institute, The Centre for Research \& Technology, Hellas, Thessaloniki, Greece (pkapoutsis@iti.gr, athakapo@iti.gr) } %{\tt\small athakapo@iti.gr}
%  %{\tt\small dkoutras@certh.gr}}%
%  \thanks{Digital Object Identifier (DOI): see top of this page.}
 
% \author{} %% Author name

\affiliation{organization={Department of Electrical and Computer Engineering, Democritus University of Thrace},%Department and Organization
            addressline={Building A, University Campus Kimmeria}, 
            city={Xanthi},
            postcode={67100}, 
            country={Greece}}

\affiliation{organization={Information Technologies Institute, The Centre for Research \& Technology, Hellas Thessaloniki, Greece},%Department and Organization
            addressline={6th km Harilaou - Thermi}, 
            city={Thessaloniki},
            postcode={57001}, 
            country={Greece}}

%% Abstract

\begin{abstract}
Unmanned Aerial Vehicles (UAVs) have revolutionized inspection tasks by offering a safer, more efficient, and flexible alternative to traditional methods. However, battery limitations often constrain their effectiveness, necessitating the development of optimized flight paths and data collection techniques. While existing approaches like coverage path planning (CPP) ensure comprehensive data collection, they can be inefficient, especially when inspecting multiple non-connected Regions of Interest (ROIs). This paper introduces the Fast Inspection of Scattered Regions (FISR) problem and proposes a novel solution, the multi-UAV Disjoint Areas Inspection (mUDAI) method. The introduced approach implements a two-fold optimization procedure, for calculating the best image capturing positions and the most efficient UAV trajectories, balancing data resolution and operational time, minimizing redundant data collection and resource consumption. The mUDAI method is designed to enable rapid, efficient inspections of scattered ROIs, making it ideal for applications such as security infrastructure assessments, agricultural inspections, and emergency site evaluations. A combination of simulated evaluations and real-world deployments is used to validate and quantify the method’s ability to improve operational efficiency while preserving high-quality data capture, demonstrating its effectiveness in real-world operations. An open-source Python implementation of the mUDAI method can be found on GitHub (\href{https://github.com/soc12/mUDAI}{\nolinkurl{github.com/soc12/mUDAI}}) and the collected and processed data from the real-world experiments are all hosted on Zenodo (\href{https://zenodo.org/records/13866483}{\nolinkurl{zenodo.org/records/13866483}}). Finally, this on-line platform (\href{https://sites.google.com/view/mudai-platform/}{\nolinkurl{sites.google.com/view/mudai-platform/}}) allows the interested readers to interact with the mUDAI method and generate their own multi-UAV FISR missions.

\end{abstract}

%\item Definition of the Fast Inspection of Scattered Regions (FISR) problem for scattered Regions of Interest (ROIs) using UAVs
%\item Multi-UAV Disjoint Areas Inspection (mUDAI) method for efficient UAV flight paths and image capturing
%\item Optimization of redundant data collection and energy-efficient UAV operations
%\item Comprehensive data quality with single-image capture per ROI
%\item Validation of efficiency and quality through real-world experiments
%\item Provision of Python implementation experimental data and interactive mission generation platform

%% Keywords
\end{frontmatter}



\section{Introduction}
\label{sec1}
%% Labels are used to cross-reference an item using \ref command.

\begin{figure*}[!ht]
    {\centering
    \includegraphics[width=1\linewidth]{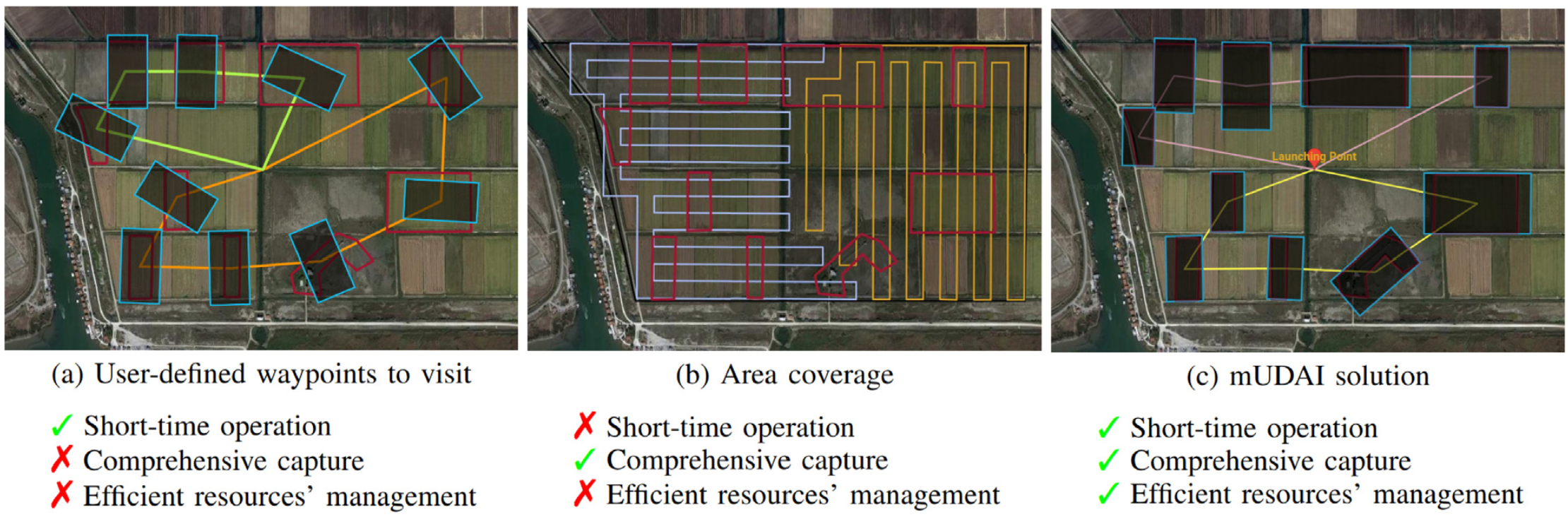}%
    \caption{Fast Inspection of Scattered Regions task. Two UAVs tasked with inspecting 10 disjoint regions: (a) User-defined waypoints allow fast operation but lack path optimization and fair task assignment, while also leading to random ROI captures. (b) CPP planning ensures high-quality captures but introduces superfluous data and long-duration operations. (c) Our mUDAI method balances speed, comprehensiveness, and efficiency, overcoming these hurdles effectively.}
    \label{fig:motivation}}
\end{figure*}

Unmanned Aerial Vehicles (UAVs) have gained prominence for their agility, adaptability, and efficiency in inspection tasks where they excel over traditional methods such as manual checks, fixed cameras, manned aircraft, and satellite imagery. These conventional approaches often require significant time, labor, and can several times pose safety risks. Their ability to reach inaccessible areas, capture high-resolution data, and operate safely and cost-effectively makes them invaluable in fields ranging from infrastructure maintenance to environmental monitoring \cite{shakhatreh2019unmanned, alzahrani2020uav}.

Despite advancements in UAV hardware, limitations, especially in battery capacity, continue to restrict both the complexity and duration of operations. To overcome these challenges, attention has shifted toward software solutions, particularly path-planning and data-gathering algorithms \cite{krestenitis2024overcome, apostolidis2022cooperative}. Such software optimizations help extend flight time, ensure high-quality data collection, and push the boundaries of UAV capabilities, enabling longer, more efficient, and more precise missions \cite{stefanopoulou2024improving}.

The problem of capturing data utilizing UAVs falls under the more generic Remote Sensing (RS) problem, and specifically in the Low-level RS category \cite{dunbar2022remote}. A particularly widespread task is the collection of data from specific Regions of Interest (ROIs), usually images, that are getting post-processed to generate high resolution representations of the scanned region(s).  The research community has dedicated significant effort towards the development and refinement of techniques known as coverage path planning (CPP) \cite{galceran2013survey} to accomplish this objective efficiently. These algorithms are designed to methodically scan entire ROIs in a standardized way, respecting user-defined parameters, to gather data. While CPP algorithms have demonstrated remarkable effectiveness in ensuring comprehensive data collection, they come with inherent drawbacks - namely, reduced speed and the potential for collecting superfluous data when a simpler capture might suffice - and they do by no means constitute a universal solution for automated UAVs' RS operations.

A certain type of operations requires to inspect (i) multiple non-connected regions, (ii) in the shortest amount of time possible, and (iii) the resolution of the collected data is important, but not critical, given that they satisfy certain specifications. Such applications could include security infrastructure inspections, agricultural inspections for certain policies (i.e., EU-CAP \cite{santiago2020comon}) or receiving funds, and emergency site assessments. While satellite imagery could fulfill some of these needs, major limitations exist, including (i) the very-low resolution (compared to UAVs' data), (ii) the inability to collect data on-demand or in real-time, and (iii) the high data acquisition and processing cost.

In the context of this work, a new problem regarding the data gathering utilizing UAVs, the Fast Inspection of Scattered Regions (FISR) is identified and defined in detail, and a novel method - multi-UAV Disjoint Areas Inspection (mUDAI) - that fulfills the objectives of this problem is proposed. The main goal of the proposed work is to balance the trade-off between the resolution of the captured data, and the time needed for the UAV(s) to perform this task. To achieve that, a methodology that allows the UAV(s) to capture a single, yet comprehensive, photo of each ROI is proposed, and trajectories that allow the fast and efficient visitation of all the defined regions are generated, in a way that (i) they respect the operational capabilities and limitations of the UAVs participating in the mission, and (ii) they facilitate in the saving of resources such as operational time and energy consumed during the procedure. Figure \ref{fig:motivation} presents an indicative example of an FISR task. The red polygons correspond to the ROIs that need to be inspected, the blue rectangles correspond to the area captured by the UAVs' sensors, and the colorful lines correspond to the trajectories that the UAV(s) should follow. In \ref{fig:motivation}-(a), the capturing positions, flight's altitude, and visitation order are user-defined parameters, leading to incomplete, or unnecessary surrounding areas' coverage, and inefficient management of operational resources (i.e., time and energy). In \ref{fig:motivation}-(b), all ROIs are covered completely, in a uniform way, however, resources are wasted to cover huge areas outside the designated regions. Finally, \ref{fig:motivation}-(c) presents the mUDAI method's solution, providing balanced length of trajectories, and the highest resolution per ROI, that ensures complete coverage.

The rest of this work is organized as follows: Section \ref{sec:relWork} presents a short literature overview of the related works, Section \ref{sec:probDef} strictly defines the FISR problem, Section \ref{sec:Method} presents the introduced methodology - mUDAI -, Section \ref{sec:sim} includes a thorough simulated evaluation of the proposed method, Section \ref{sec:real} presents two sets of real-world deployments along with evaluation and comparison results, and finally, Section \ref{sec:Conclusions} summarizes this work and discusses our thoughts and future plans.

%% Use \subsection commands to start a subsection.

\section{Related Work}
\label{sec:relWork}
To the best of our knowledge, at the moment this work is written, there are no works available in literature with the objective to capture a single representative image of multiple disjoint ROIs during a single UAV flight. The most common approach used in such cases is CPP, where (i) either scanning a wider region - including all the sub-ROIs - to obtain the desired data is involved, leading though to waste of resources (i.e., time and energy) to also scan regions that are of no interest, or (ii) a combination of CPP and the Traveling Salesman Problem (TSP) \cite{flood1956traveling} is applied in disjoint ROIs, scanning each of them separately, and determining the visitation order using a TSP solution to make the procedure more efficient. \cite{almadhoun2019survey, cabreira2019survey} presents an overview of the recent CPP works for UAVs' coverage operations, while \cite{thibbotuwawa2020unmanned} presents a literature review on VRP \cite{toth2002vehicle} - generalization of the TSP, utilizing multiple vehicles - applications for UAVs. While neither CPP, nor TSP-CPP (as usually met in literature) methods can be considered as directly related to the work introduced in this paper, these two, along with human-defined points of interest (POIs) to be visited by a UAV, are the approaches that can be used as alternatives to fulfill (not optimally) the objectives of the FISR problem. Below are shortly presented some recent, interesting works of these two fields.

\cite{ghaddar2020pps} presents an ``energy aware'' CPP method for multiple UAVs, using boustrophedon \cite{choset1998coverage} - a typical cellular decomposition method combined with the back-and-forth scanning pattern - as the core CPP method. This work supports the presence of with no-fly-zones (NFZs) inside the ROIs, and the efficiency of the proposed methodology is evaluated via simulations. \cite{luna2022fast} introduces a methodology for efficient real-world coverage operations with multiple UAVs. Similar to \cite{ghaddar2020pps} it uses boustrophedon as the core CPP method, and it demonstrates the efficiency of the overall proposed methodology in real-world operations using custom developed UAVs. \cite{apostolidis2022cooperative} introduces an end-to-end coverage mission planning platform for multiple UAVs, that focuses on the efficient deployment of grid-based CPP methodologies in real-world operations, being able to handle very complex-shaped ROIs NFZs inside them. This work, that uses Spanning Tree Coverage (STC) \cite{gabriely2001spanning} as the core CPP method, extensively evaluates the proposed coverage scheme in simulated runs, and demonstrates the platform's efficiency on different real-world scenarios with commercial UAVs. Finally, \cite{apostolidis2023systematically} extends the path planning methodology introduced in \cite{apostolidis2022cooperative} in a way that it further increases the efficiency of grid-based methods in real-world operations, introducing three ad-hoc coverage modes that fulfill objectives appropriate for different types of coverage operations. The efficiency of the coverage modes is evaluated in simulated runs.

\cite{xie2020path} deals with TSP-CPP, providing a mixed integer programming formulation for the TSP problem at first, and dealing with the coverage of the separate ROIs as the next step. This work uses boustrophedon as the core CPP method, and the evaluation of the method is performed in simulations. \cite{xie2022multiregional} also deals with the problem of CPP in disjoint ROIs, however, in this case utilizing multiple UAVs for the coverage procedure. The authors approach it as a mix of multiple TSP (mTSP) and CPP. Here, as in \cite{xie2020path} boustrophedon is used as the core CPP method, and the efficiency of the proposed scheme is evaluated via simulations. Finally, in \cite{luna2023spiral} the authors approach the same problem in a slightly different way. Specifically, a scheme is proposed where the spiral pattern CPP is used as the core method, equal task assignment for the UAVs is performed, where each task may involve more than one ROIs, and a TSP problem is finally solved to find the most efficient order for the UAVs to visit the disjoint regions. The proposed method is evaluated in simulations. In a complementary vein, swarm-based inspection studies have emerged, such as \cite{gao2024hierarchical}, which employs hierarchical partitioning with viewpoint generation and route planning for each UAV, and \cite{liu2025advancing}, which introduces a bi-level task assignment and multi-objective routing scheme. Although designed for multi-UAV settings, these works formalize the selection of tasks/regions and the sequencing of visits.

%% Use \subsubsection, \paragraph, \subparagraph commands to 
%% start 3rd, 4th and 5th level sections.
%% Refer following link for more details.
%% https://en.wikibooks.org/wiki/LaTeX/Document_Structure#Sectioning_commands
\section{Problem Definition: Fast Inspection of Scattered Regions (FISR)}
\label{sec:probDef}
This section formulates the problem of \textit{Fast Inspection of Scattered Regions (FISR)}, i.e., collecting the best possible images from spatially disjoint ROIs in the minimum possible time utilizing UAVs.

\subsection{The Setup}

Assuming a user-defined set of $k$ ROIs that should be inspected:
\begin{equation}
\label{eq:rois}
    {\cal R} = \{R_1,\dots,R_k\}
\end{equation} 

Each ROI $R_i$ is a simple polygon in a 2‑D Cartesian reference frame (east–north; meters), $R_i = [r_1, r_2,\dots, r_{p_i}], \; r_\ell  = [x_\ell, y_\ell]^\top \in \mathbb{R}^2$ with vertices ordered counter‑clockwise and $r_{p_i} \equiv r_1$. The integer $p_i$ denotes the number of vertices\footnote{Please note that any other coordinates' system is also compatible with such a definition and a change of reference frame is handled by a known rigid transformation (rotation and translation).}. 

\subsection{Decision Variables}
Assuming a team of $n$ UAVs capable of executing waypoint-based trajectories, the decision variables for the whole swarm are defined by 

\begin{equation}
\label{eq:controllable_variables}
\tau :=  [ \tau_1, \tau_2, \dots, \tau_{n} ]
\end{equation}

where $\tau_i, \; \forall i \in \{1,2, \dots, n\}$ denotes the trajectory of the $i$th UAV and is defined as follows:

\begin{equation}
\label{eq:robotDecisions}
\tau_i = [v^i_1, v^i_2, \dots, v^i_{m_i}],
\end{equation}

where each ``viewpoint'' of the above trajectory is defined by
\begin{equation}
\label{eq:viewpoint}
v^i_j = [x, y, z, \psi]^\top \in \mathbb{R}^4, \; \forall i \in \{1, 2, \dots,n \}, \; \forall j\in \{1, 2, \dots, m_i \},
\end{equation}

where $x$ and $y$ denote the 2-D Cartesian coordinates, $z$ the altitude relative to the ground and $\psi$ is the orientation (yaw) of the UAV\footnote{Although we have used the usual assumption for this type of missions that the gimbal pitch is fixed and we control only the image orientation via yaw, the formulation readily extends to additional sensor controls available to each UAV payload.}.

For each $i$th UAV, the definition of all viewpoints (\ref{eq:viewpoint}) cannot be set arbitrarily, but they need to satisfy a set of operational constraints.  More specifically: 
\begin{enumerate}
    \item \textbf{Battery Limitation}. $\mathcal{C}_1(\tau_i) := \mathrm{dur}(\tau_i)-t_{\max} \leq  0,\: \forall i \in \{1,2,\dots,n\}$, where $\mbox{dur}(\tau_i)$ denotes the time needed to execute $\tau_i$ trajectory and  $t_{max}$ is a maximum operational duration - in line with the battery capabilities of each $i$th UAV.
    \item \textbf{Maximum speed}. $ \mathcal{C}_2(v^i_j,v^i_{j+1}) := \bigl\lVert \pi_{xyz}(v^i_{j+1})-\pi_{xyz}(v^i_j)\bigr\rVert_2 - u_{\max}\,\Delta t \leq 0$, $\forall i \in \{1,2,\dots,n\}$, $j \in \{1,2, \dots, m_i-1\}$, where $\pi_{xyz} : \mathbb{R}^4 \to \mathbb{R}^3$ denotes a projection that discards $\psi$ variable, i.e., $\pi_{xyz}(x,y,z,\psi) = (x,y,z)$, $u_{max}$ denotes the maximum speed the UAVs are allowed to travel and $\Delta t$ the control-execution loop time. 
    \item \textbf{Altitude within range}. $\mathcal{C}_3(v^i_j) := \max\{a_{min}-z^i_j,\ z^i_j-a_{max}\} \leq  0,\:  \forall i \in \{1,2,\dots,n\},\: j \in \{1,2,\dots, m_i\}$ where $a_{min},\: a_{max}$ denote the minimum and maximum altitude a UAV is allowed to reach (due to hardware limits, user-defined settings, or regulations) respectively.
    \item\textbf{Sensor Specification}. $\mathcal{C}_4(v^i_j) := \text{sensor-spec constraint at }v^i_j$. The UAVs carry sensors with certain specifications (i.e., horizontal field of view (hFOV), vertical field of view (vFOV), horizontal resolution (hRes), vertical resolution (vRes), etc. \cite{santos2014handbook}).
\end{enumerate}
For notation purposes, all the previously defined constraints $\mathcal{C} \equiv \big(\mathcal{C}_1,\mathcal{C}_2,\mathcal{C}_3,\mathcal{C}_4\big)$ can be, in general, represented as a system of inequalities:
\begin{equation}
    \label{eq:constraints}
    {\cal C}(\tau) \leq 0
\end{equation}
where ${\cal C}$ is a set of nonlinear constraint functions of the decision variables $\tau$. 

\subsection{Measurements - Collected Images}
After reaching each viewpoint  $v^i_j$, an image $I_j^i \in \mbox{Img}$ (where $\mbox{Img}$ denotes the space of images), $\forall i \in \{1, 2, \dots,n \}$, $\forall j\in \{1, 2, \dots, m_i \}$  is captured, forming the measurements' vector:
$$
y_i = [I_1^i, I_2^i, \dots, I_{m_i}^i],\: \forall i \in \{1,2,\dots,n\}
$$
The aggregate measurement collection for the mission is:
\begin{equation}
\label{eq:data_vec}
y := [y_1, y_2, \dots, y_n]
\end{equation}
\subsection{Objective Function}
%The objective of the FISR problem is to acquire the best possible representation of the user-defined set of $k$ ROIs (\ref{eq:rois}). 
After the collection of image data (\ref{eq:data_vec}), a set of ROIs - Images ordered pairs can be formed:
\begin{equation}
\label{eq:q}
Q_i(y) = (R_i, {\cal I}(R_i)),\: \forall i \in \{1,2,\dots,k\}
\end{equation}
where ${\cal I}(R_i)$ denotes the corresponding image for $R_i$. For each $k$ pair, $J(Q_i): \text{(ROI,image)} \mapsto \mathbb{R}_{\ge 0}$ denotes a score that quantifies the \textit{completeness} and \textit{accuracy}\footnote{You should take into consideration, that the definition of $J(\cdot)$ is strictly application-dependent; however, in the following section, we provide 2 concrete instantiations of this function, capturing different needs of the FISR application.} of representation of $R_i$ from ${\cal I}(R_i)$. It is assumed that $J(Q_i)=0$ means no alignment at all and the objective is to maximize such score, therefore, $J \in [0, \infty)$. The overall representation score can be defined as:
\begin{equation}
\label{eq:score}
{\cal J}(y) = \sum_{i=1}^k J(Q_i(y))
\end{equation}

All in all, the objective of such a problem is to be able to achieve the best possible score of (\ref{eq:score}) in the minimum possible time. The naive definition of the objective function would have to linearly combine the representation score of (\ref{eq:score}) with the time needed, e.g., ${\cal L}(y, \tau) = {\cal J}(y) - \alpha \sum_{i=0}^n \mbox{dur}(\tau_i)$, where $\alpha$ is a positive constant that scales the two terms, also weighting their importance. However, choosing $\alpha$ is application-dependent and non-trivial. On top of that, other non-linear combinations of the two factors may be beneficial, making the loss function problem even more intractable. Finally, in the majority of the real-world applications, it is usually preferred to prioritize the acquisition of the best possible captures (\ref{eq:data_vec}) of (\ref{eq:rois}) and then, with lower priority, to find the minimum paths (\ref{eq:controllable_variables}) that guide the UAVs to these capture positions.

\subsection{FISR Optimization Problem}
The problem of FISR could be reduced by the observation that you could first calculate the best possible viewpoints $v^*$ independently of the number of UAVs, trajectories, etc. Thus, for each $R_i, \; \forall i \in \{1, 2, \dots,k \}$ (\ref{eq:rois}), the best possible viewpoint $v_i^*$ can be computed by solving the following constrained (\ref{eq:constraints}) optimization problem:
\begin{equation}
    \label{eq:optimizeViewpoints}
    \begin{array}{l}
    v^*_i = \argmax_{v_i} J(Q_i(y)) \\
    \mbox{subject to }\; \mathcal{C}_{\{3,4\}}(v_i)\le 0
    \end{array}
\end{equation}

Now, having the optimal set of viewpoints, the UAV trajectories design problem for the FISR problem gets reduced to minimizing the travel time of the UAV with the longer trajectory (\ref{eq:robotDecisions}), forcing that these exact optimal viewpoints of (\ref{eq:optimizeViewpoints}) will be contained in the UAV trajectories and all the (\ref{eq:controllable_variables}) operational constraints are met:
\begin{equation}
    \label{eq:optimizeTime}
    \begin{array}{rl}
	\mbox{minimize} & \max_{i \in \{1, 2, \dots, n\}}\mbox{dur}(\tau_i)\\
	\mbox{subject to} & \bigcup_{i=1}^n \{\,v^i_j : j=1,\dots,m_i\,\} \;\supseteq\; \{\,v^*_1,\dots,v^*_k\,\},\\
                      &\mathcal{C}_1(\tau_i)\le 0,\quad \forall i,\\
&\mathcal{C}_2(v^i_j,v^i_{j+1})\le 0,\quad\forall i,\ \forall j=1,\dots,m_i-1,\\
&\mathcal{C}_3(v^i_j)\le 0,\quad \forall i,\ \forall j=1,\dots,m_i.
	\end{array}
\end{equation}

Such a transformation significantly reduces the complexity of the whole problem, while at the same time, allows for application-dependent tackling of each optimization setup, since (\ref{eq:optimizeTime}) is completely agnostic to how the viewpoints are defined.

\begin{figure*}[!h]
    {\centering
    \includegraphics[width=0.9\linewidth]{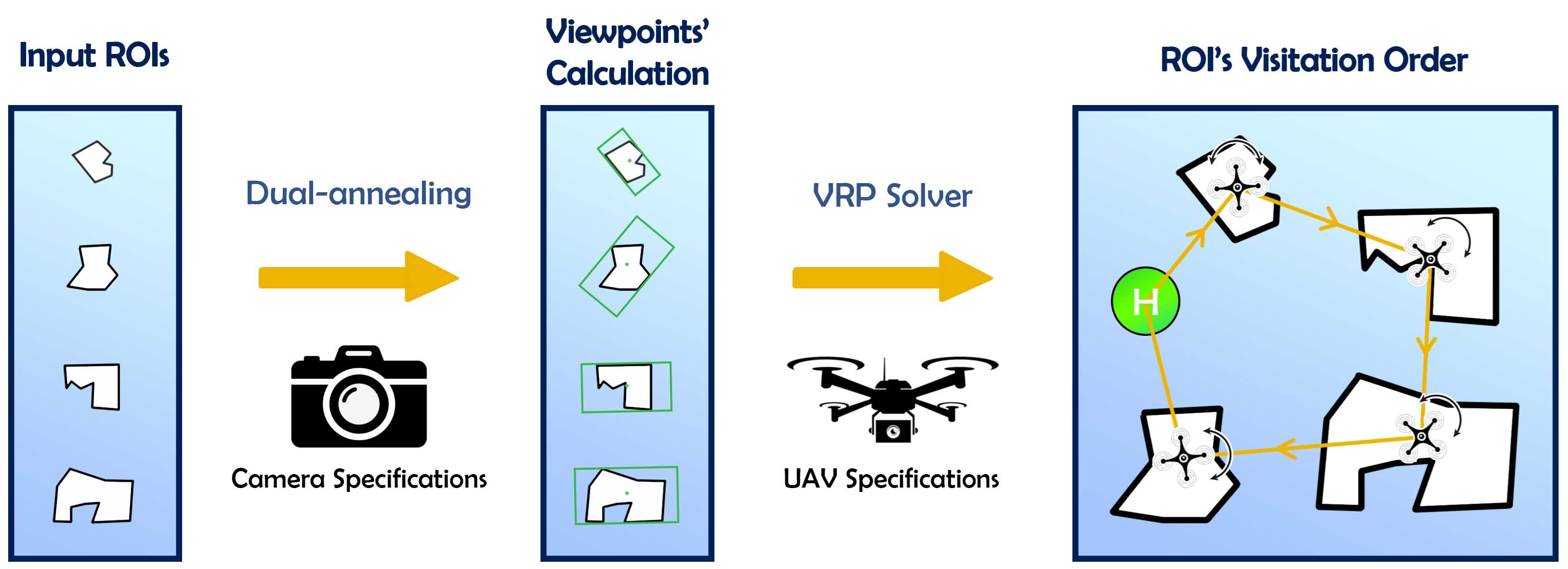}%
    \caption{mUDAI methodology splits the FISR problem in two discrete optimization sub-problems - as a first step, given the ROIs and the specifications of the camera used, dual-annealing algorithm is utilized for the calculation of the optimal viewpoints, given the optimization objectives - as a second step, a VRP solver is used for the calculation of the optimal visitation order, respecting the energy constraints of the UAVs.}
    \label{fig:methodology}}
\end{figure*}

\section{multi-UAV Disjoint Areas Inspection (mUDAI) Methodology}
\label{sec:Method}

\subsection{Overview of the approach}

Consistent with Section~\ref{sec:probDef}, our proposed \textit{mUDAI} methodology follows a two-stage, viewpoint-first strategy: 
(i) \textbf{Viewpoint Optimization} (\ref{eq:optimizeViewpoints}) – determine for each ROI $R_i$ the viewpoint $v_i^*$ that maximizes representation quality $J(Q_i(y)$, ensuring that the captured image $I_j^i$ provides a high-quality depiction ${\cal I}(R_i)$; 
(ii) \textbf{Trajectory Optimization} (\ref{eq:optimizeTime}) – plan UAV trajectories $\tau$ that connect initial positions to the selected viewpoints and then to the designated endpoint, while satisfying ${\cal C}(\tau)\!\le\!0$ and minimizing mission duration. A schematic overview of the mUDAI's pipeline is depicted in Figure \ref{fig:methodology}.

% Our proposed \textit{mUDAI} methodology addresses this problem in a structured, two-stage approach:
% \begin{enumerate}
%     \item Viewpoint Optimization:
%     We first focus on selecting a single optimal viewpoint for each ROI, ensuring that the image $I_j^i$ acquired at this viewpoint provides a high-quality representation ${\cal I}(R_i)$ for the corresponding ROI $R_i$. 
% \item Trajectory Optimization: Once the optimal viewpoints are computed, the second step involves the optimization of trajectories $\tau$ that connect the UAVs' initial positions to the selected viewpoints $v_i$, and then back to a designated endpoint.
% \end{enumerate}
% Our approach effectively combines two distinct optimization strategies: one for determining the best photo capture positions and another for planning optimal UAV flight routes. This two-step setup provides both precise data capture and efficient navigation. A schematic overview of the mUDAI's pipeline is depicted in Figure \ref{fig:methodology}.

\subsection{Viewpoint Optimization}
The primary goal of the mUDAI algorithm is to optimize the capture process while considering flight constraints. By focusing on a single viewpoint per ROI, the complexity of selecting multiple overlapping or redundant viewpoints is reduced. To find these optimal viewpoints, we leverage the UAV's camera model and the ROI's geometric representation. Figure \ref{fig:CC} visualizes the projection of the region $R_i$ on the captured image, where $I_j^i$ corresponds to a box of $d_h$ length and $d_v$ width:

\begin{figure}[!b]
    \centering
    \includegraphics[width=.98\linewidth]{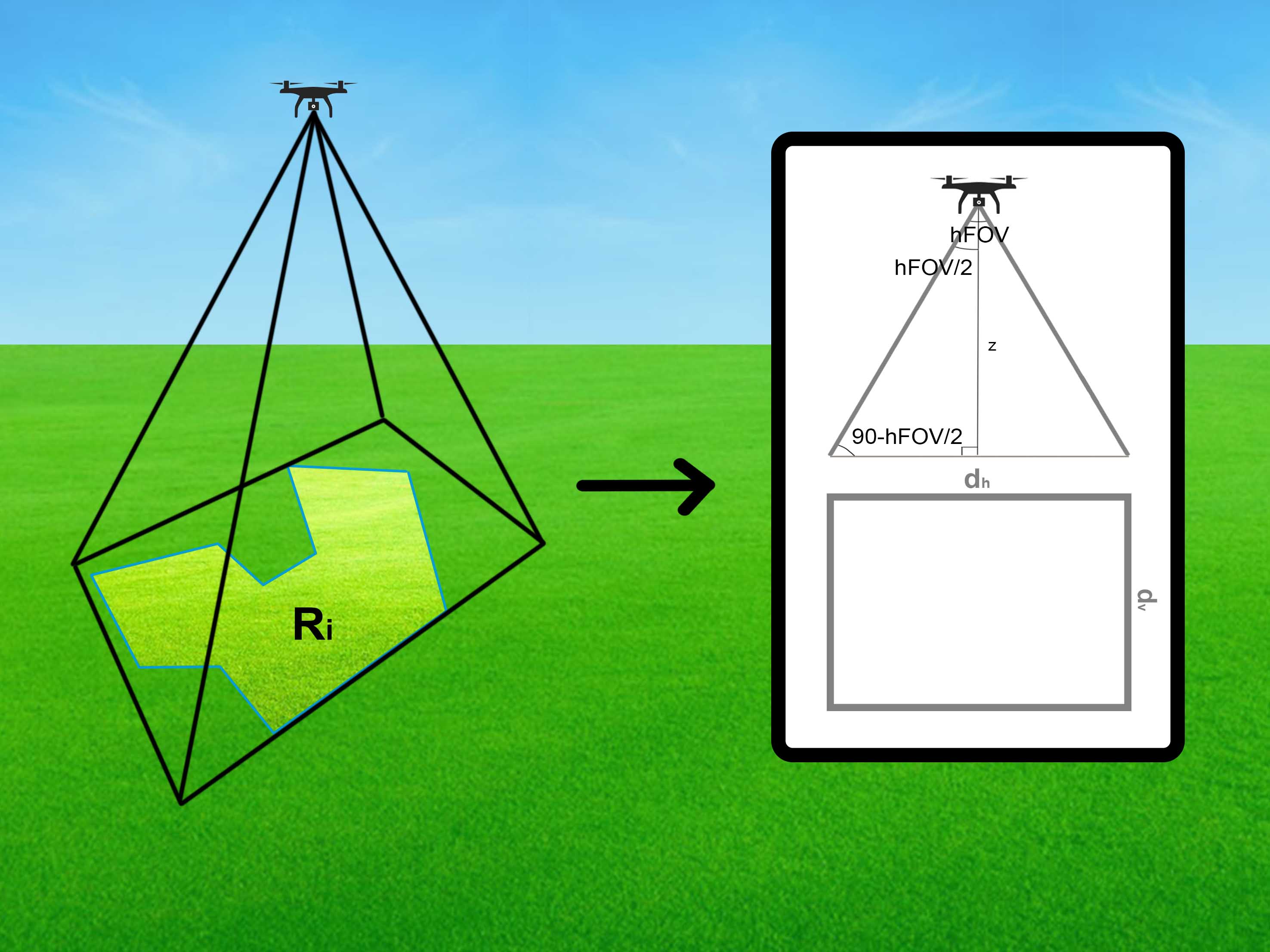}%
    \caption{The UAV, hovering in a distance from ground -z- captures an image that contains a representative shot of the region $R_i$. The aspect ratio of the captured image $(d_h:d_v)$ depends on the specifications of the sensor (hFOV, vFOV), while the content of the captured image -given the aspect ratio- depends on the position and rotation of the UAV $[x, y, z, \psi]$.}
    \label{fig:CC}
\end{figure}

\begin{equation}
	\label{eq:dh_covered}
	d_h = 2 \ \frac{h \ \sin(HFOV/2)}{\sin(90-HFOV/2)} \Rightarrow
	d_h = 2 \ h \ \tan(HFOV/2)
\end{equation}

\begin{equation}
	\label{eq:dv_covered}
	d_v = 2 \ \frac{h \ \sin(VFOV/2)}{\sin(90-VFOV/2)} \Rightarrow
	d_v = 2 \ h \ \tan(VFOV/2)
\end{equation}
More formally, for each ROI $R_i$, our objective is to determine a viewing pose $v^i_j = [x,y,z,\psi]$ that maximizes $J(Q_i(\textbf{y}))$  during scene imaging and specifically to optimize (\ref{eq:optimizeViewpoints}).

To achieve this, we consider two possible optimization metrics, which can be directly substituted into $J(Q_i)$ (\ref{eq:score}):

\begin{enumerate}   
    \item \underline{Maximized Coverage Objective (MCO)}:
    Ensures each ROI is completely captured in a single photo, minimizing the inclusion of areas outside the intended target.
    \item \underline{Balanced Coverage Objective (BCO)}:  
    Seeks a compromise between capturing the entire ROI and reducing the capture of extraneous space, focusing instead on high-resolution captures of significant ROI’s portions.
\end{enumerate}

The first objective, denoted as $MCO$, prioritizes maximizing the overlap between the ROI and the camera capture. When this overlap is maximized, indicating that the entire ROI is contained within the camera capture, an additional penalty term is introduced to penalize large captures. This penalty is applied only when the whole ROI is contained within the capture, ensuring that the capture is as small as possible while still fully covering the ROI. $MCO$ is computed, over all collected images (\ref{eq:data_vec}), using the formula
\begin{equation}
    {\cal J}_{\textbf{MCO}}(y) = 
    \begin{cases} 
\frac{R_i \cap \mathcal{I}(R_i)}{R_i}  & \text{if } \frac{R_i \cap \mathcal{I}(R_i)}{R_i} < 1  \\
\frac{R_i \cap \mathcal{I}(R_i)}{R_i} + \frac{1}{\mathcal{I}(R_i)} & \text{otherwise }  \left(\mbox{} \frac{R_i \cap \mathcal{I}(R_i)}{R_i} = 1 \right)
\end{cases}
\end{equation}

The second objective, $BCO$, assigns equal weight to both the intersection and union of the target ROI and camera capture, resembling the popular Intersection over Union (IoU) metric used in various applications. To calculate $BCO$ we apply the formula
\begin{equation}
    {\cal{J}}_{\textbf{BCO}}(y) =
    \frac{R_i \cap {\cal I}(R_i)}{R_i \cup {\cal I}(R_i)}
\end{equation}

\begin{figure}[ht]
	\centering
	\includegraphics[width=0.98\linewidth]{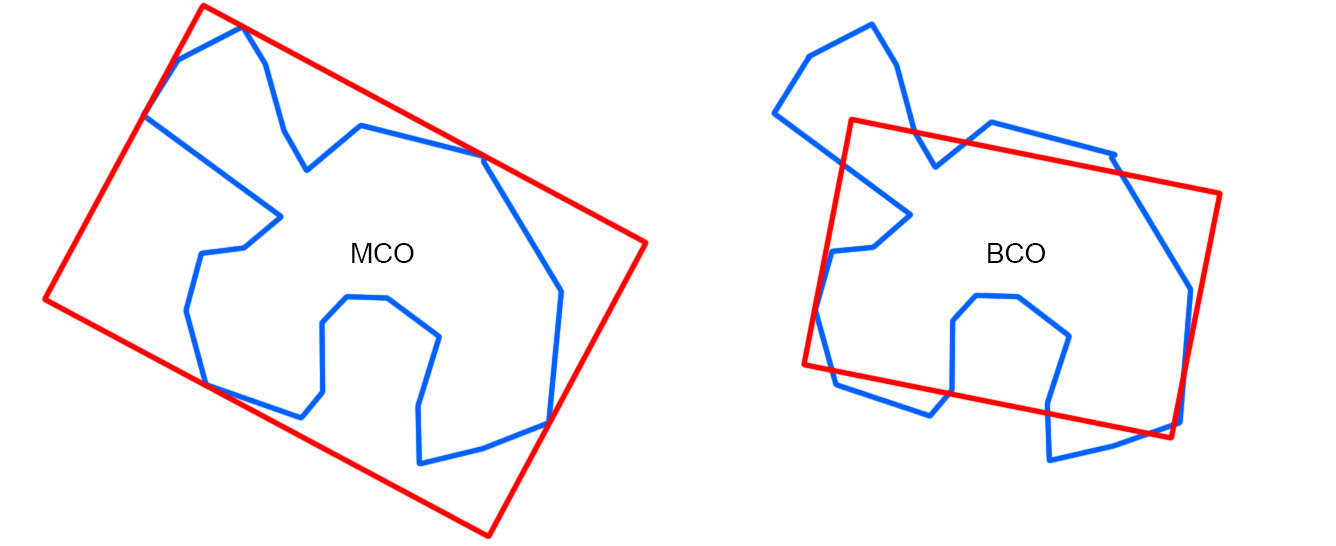}
	\caption{Illustrative comparison of MCO and BCO optimization objectives. The blue polygon represents the ROI, while the red polygon indicates the camera capture area. In MCO, the objective is to fully capture the defined ROI while minimizing the area outside the ROI. In contrast, BCO strikes a balance between capturing the ROI and reducing coverage of areas outside the ROI, achieving a lower GSD and minimizing extraneous coverage.}
    \label{fig:optim}
\end{figure}

Figure \ref{fig:optim} showcases the results of the two objectives for the same ROI. By considering these objectives, mUDAI can capture the entire ROI, or focus on obtaining high-resolution imagery of the most informative areas, while minimizing the impact of any exterior noise. This adaptability allows for efficient and effective capture planning in various scenarios.

For each $R_i$, we solve a different optimization problem using the following optimization objectives. For the optimization procedure, we employ the dual-annealing algorithm. Dual-annealing operates by iteratively adjusting the position and orientation of the UAV in each direction, which includes longitude (\(x\)), latitude (\(y\)), altitude (\(z\)), and yaw angle (\(\psi\)) to achieve the best photo capture.

The inputs consist of: 1) an Objective Function, which is the function to be optimized by taking a set of input parameters $[x, y, z, \psi]$ and returning the objective value; 2) an Initial Solution $[x_0, y_0, z_0, \psi_0]$, which serves as the starting point for the optimization process; and 3) Bounds for each input parameter $[x_b, y_b, z_b, \psi_b]$, which are to limit the permissible range of values for these parameters. The bounds are defined as follows:
\begin{itemize}
    \item $x_b$: The range of allowable values for the x-coordinate of the UAV's position. It is determined by the $[x_{\text{min}}, x_{\text{max}}]$ of the ROI's polygon.
    \item $y_b$: The range of allowable values for the y-coordinate of the UAV's position. It is determined by the $[y_{\text{min}}, y_{\text{max}}]$ of the ROI's polygon.
    \item $z_b$: The range of allowable values for the altitude of the UAV. It is bounded between the minimum altitude $[{a_{min}]}$ and the maximum altitude $[{a_{max}]}$, two user-defined parameters that may derive either from the topology of the region, or the flight regulations.
    \item $\psi_b$: The range of allowable values for the parameter psi, which represents the heading of the UAV. In our implementation it is bounded between 0 and 180 degrees.
\end{itemize}

By utilizing the dual-annealing algorithm with the appropriate inputs, the optimization process can efficiently explore the solution space and find the optimal configuration that maximizes the $J(Q_i)$ score based on the defined constraints and bounds (part of ${\cal C}(\tau)$).

\subsection{Trajectory Optimization}

VRP is a fundamental optimization problem in logistics and transportation. In VRP, a fleet of vehicles is tasked with delivering goods or servicing various locations. Each location has its own demand or service requirement, and the objective is to find the most efficient routes for the vehicles to fulfill all demands while optimizing factors such as distance traveled and vehicle utilization.

In our case, the VRP problem is time-constrained, as the battery capacity constraint directly translates into the number of ROIs that a UAV can capture in a single flight. The UAV's battery capacity limits the distance it can cover and the duration it can stay operational during the inspection mission. Therefore, the routing problem in FISR involves finding the optimal paths for the UAVs to inspect the scattered regions or parcels while ensuring that the battery capacity is not exceeded.

More precisely, this step determines the order and allocation of ROIs to UAVs, aiming to solve (\ref{eq:optimizeTime}). By integrating the results from the viewpoint optimization stage, the VRP ensures each $R_i$ is inspected at its chosen viewpoint $v^i_j$. The resulting set of trajectories $\tau_i$ respects the maximum speed, altitude, and time constraints and positions the UAVs to capture high-quality images effectively.

To solve this problem, we used Google OR-tools\footnote{\url{https://developers.google.com/optimization}}. Specifically, we configure the Routing Solver with Path Cheapest Arc as the first-solution strategy to quickly build a feasible baseline by greedily extending each UAV route with the lowest-cost available arc; we then apply Guided Local Search (GLS) with standard local-search moves to refine the routes and avoid local minima by adaptively penalizing costly or overused arcs, temporarily modifying the objective and encouraging exploration of alternative connections. This combination yields feasible multi-UAV trajectories that respect the stated distance/battery constraints while improving efficiency beyond the greedy baseline. While this configuration serves our purposes well, it is not intended as a prescription; the framework is intentionally modular and solver-agnostic, and any suitable mTSP/VRP method (heuristic or exact) can be substituted as a drop-in replacement, provided it adheres to the same constraints defined in our formulation.

\begin{figure*}[h]
	\centering
	\includegraphics[width=0.83\textwidth]{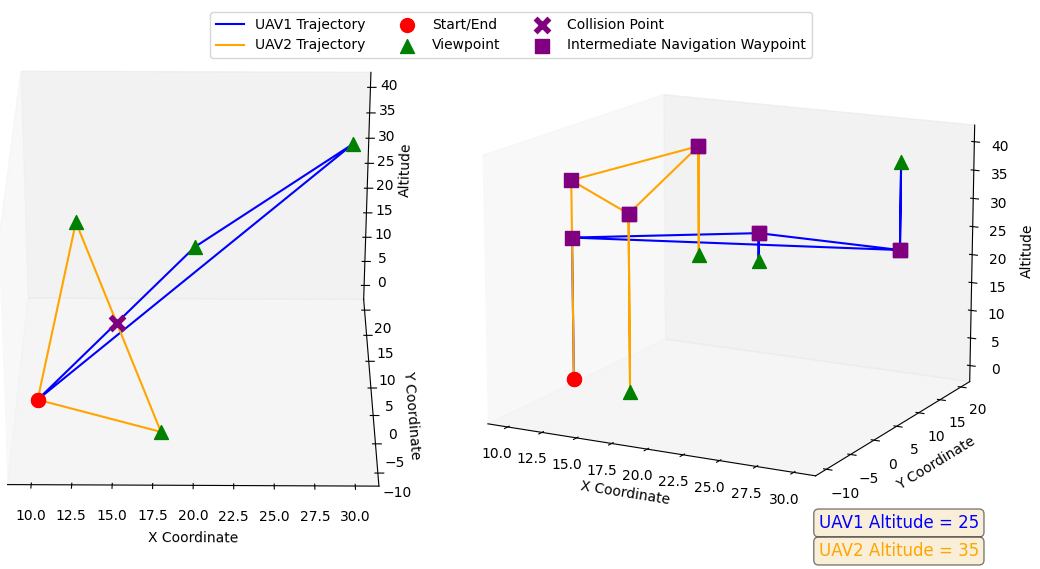}
	\caption{The inclusion of intermediate navigational waypoints in UAV trajectories, enabling transitions at constant altitudes unique to each UAV, ensures safe and collision-free operations. On the left side of the figure, a trajectory directly connecting the viewpoints is shown, which includes intersecting trajectories leading to a potential collision point, whereas on the right side, the introduction of intermediate waypoints has eliminated the presence of potential collision points.}
	\label{fig:interm_way}
\end{figure*}

% To solve this problem, we utilized the Google OR-tools\footnote{\url{https://developers.google.com/optimization}}. OR-Tools is a powerful and versatile optimization software suite designed to address complex optimization problems related to vehicle routing, flows, integer and linear programming, and constraint programming. The existing software underwent considerable modifications to align with our specific requirements, particularly addressing two critical areas: path generation and efficient UAV deployment.

Firstly, the application was adapted to incorporate constraints that factor in the maximum distance the UAV can cover. This ensures that the generated paths are within the operational capacity of the UAV. The algorithm calculates the distances between all the ROIs and the initial position, which is crucial for optimizing travel time. The initial position is selected to be as centrally located as possible among the ROIs, thereby minimizing unnecessary travel and simplifying the launch and return phases. Since the UAVs start and end their journeys at this same location, the flight paths form a closed loop.

The VRP solution is initially computed given the current number of available UAVs. If the resulting route segments exceed the UAVs’ battery capabilities, the route is then split into twice as many segments for the next attempt. In other words, a single UAV’s route is subdivided into two segments if needed; if there are three UAVs in use, their three individual routes become six upon splitting. This doubling process continues until each resulting segment fits within the UAVs’ operational battery limits, ensuring that the mission remains both feasible and efficient.

Our approach supports a multi-UAV setup that operates in parallel by assigning distinct navigation altitudes for each UAV. This prevents potential collisions when UAVs travel diagonally between viewpoints as demonstrated in Figure \ref{fig:interm_way}. Each UAV travels at a unique altitude between viewpoints, adjusts its altitude and orientation upon reaching a viewpoint to capture the ROI, and then resumes its assigned navigation altitude to travel to the next viewpoint. This process ensures safe and efficient operation, eliminating the risk of collision.

\section{Simulated Evaluations}
\label{sec:sim}
This section presents an extensive simulation-based assessment of the proposed methodology, designed to validate and quantify its claimed functionalities and effectiveness. The evaluation focuses on image capture performance, multi-UAV deployment efficiency, and critical operational indicators such as travel time, distance, and energy consumption. Section \ref{subsubsec:View} details the simulation results related to viewpoint optimization, whereas Section \ref{subsec:multi-UAV} analyzes the multi-UAV performance of the system.

\subsection{Viewpoint Optimization Study}
\label{subsubsec:View}

\begin{figure*}[!ht]
	\centering
	\includegraphics[width=0.86\textwidth]{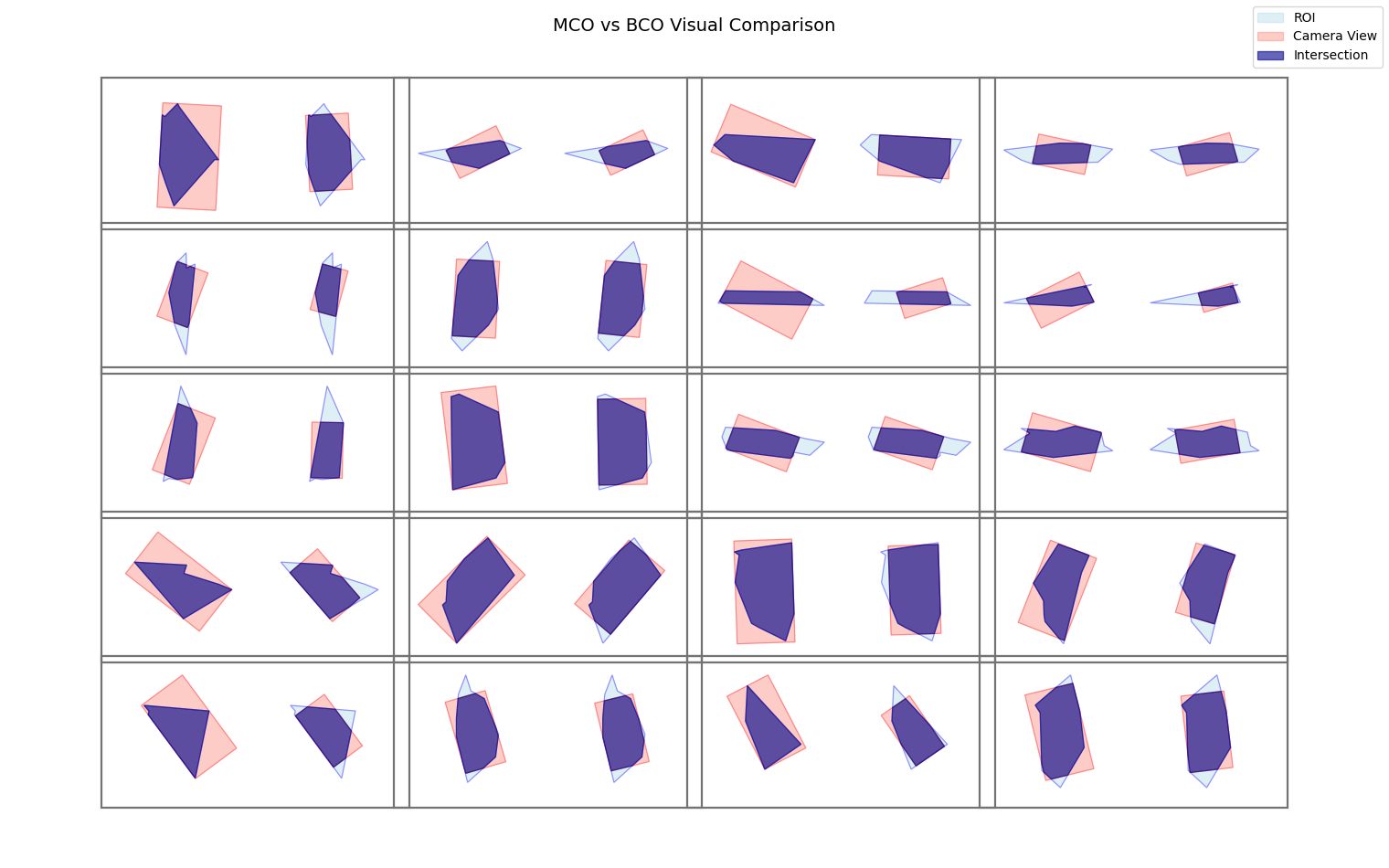}
	\caption{A visual comparison of MCO and BCO Objectives for 20 random polygons from the Simulated Evaluations. }
	\label{fig:50_poly}
\end{figure*}

This section presents a series of indicative experiments employing the BCO and MCO objective functions in order to validate our observations both qualitatively and quantitatively. For this purpose, a random polygon generator was utilized to produce a set of 50 polygons with systematic variations in size, angular configuration, and number of vertices. The dataset was further designed to include challenging instances, such as elongated polygons aligned with a principal axis, which deviate from the camera’s field-of-view specifications and resolution limits. Representative outcomes for a subset of 20 polygons are illustrated in Figure \ref{fig:50_poly}, while the aggregate quantitative results across all 50 cases are reported in Table \ref{tab:MCO_BCO_comp}, which presents the average values across five repeated experiments.

To elaborate, Figure 6 illustrates and validates our proposed solutions, each of which addresses a specific scenario. It can be observed that the MCO objective tends to capture the entire polygon whenever feasible (failing to achieve full coverage is attributed to altitude limitations), whereas the BCO objective seeks a compromise between maximizing the region of interest (ROI) and limiting the inclusion of external noise. This contrast is particularly evident in the case of elongated polygons, where the MCO, in attempting to encompass the complete ROI, inevitably includes substantial out-of-ROI regions that may be redundant or even detrimental to performance, depending on the downstream task.

To evaluate the coverage performance for each ROI in our experiments, we adopted the standard retrieval metrics of Recall and Precision. Recall quantifies how much of the target area was actually covered. It is defined as the ratio of correctly covered area (true positives) to the total area that should have been covered (true positives + false negatives). Precision, on the other hand, reflects the accuracy of the coverage, i.e., how much of the covered area was indeed relevant. It is calculated as the ratio of correctly covered area (true positives) to the total area that was covered (true positives + false positives).

\begin{table}[!ht]   
    \centering
    \resizebox{\columnwidth}{!}{%
    \begin{tabular}{c|ccccc}
         Objective & Recall & Precision & GSD & Deviation & Iterations\\
         \hline\hline
         MCO & 91.64\% & 59.88\% & 2.90 cm/pixel & 0.0005\% & 6087\\
         \hline
         BCO & 78.24\% & 80.20\% & 2.32 cm/pixel & 0.003\% & 915\\
    \end{tabular}
    }
    \caption{Comparison of the MCO and BCO objectives in the simulated experiments. Reported values correspond to the mean performance over five runs, showing recall, precision, ground sampling distance (GSD), relative deviation, and convergence iterations.}
    \label{tab:MCO_BCO_comp}
\end{table}

The quantitative results in Table \ref{tab:MCO_BCO_comp} reinforce the distinction mentioned above. Specifically, the MCO achieves a recall of 91.64\% but with a lower precision of 59.88\%, reflecting its tendency to over-include areas beyond the ROI. Conversely, the BCO achieves a more balanced performance, with a recall of 78.24\% and a precision of 80.20\%, indicating a more selective capture of the ROI while suppressing extraneous regions. As expected, BCO features better GSD as it is more focused on the ROI with 2.32 cm/pixel compared to MCO's 2.90 cm/pixel. These findings suggest that while MCO is advantageous when completeness of coverage is critical, BCO provides a more efficient trade-off between coverage and noise reduction, particularly in scenarios involving irregular or elongated geometries.

To ensure robustness of the reported values, each experiment was repeated five times, and the presented recall, precision, GSD and iterations correspond to the mean performance across runs. Furthermore, the relative deviation was computed in each case to quantify variability. The deviation depicted is negligible and should not have any meaningful effect in the Viewpoints. Finally, the iterations column reports the iteration index at which the best objective value was first observed; it equals the total number of evaluations across the global and local phases of dual-annealing (i.e., their sum). Lower values therefore indicate faster convergence. Unsurprisingly, BCO converges much faster, 915 iterations on average, whereas MCO requires 6,087 on average. The BCO objective is a smooth Jaccard ratio with gradual variation, yielding broad attraction basins and faster convergence. In contrast, the MCO objective is piecewise, with a flat plateau below full coverage and a narrow ridge at exact coverage where the secondary term activates, creating a larger and more complex search space. This explains why MCO requires significantly more iterations to converge.

\subsection{Multi-UAV Study}
\label{subsec:multi-UAV}
\begin{figure*}[!ht]
    \centering
	\includegraphics[width=.7\linewidth]{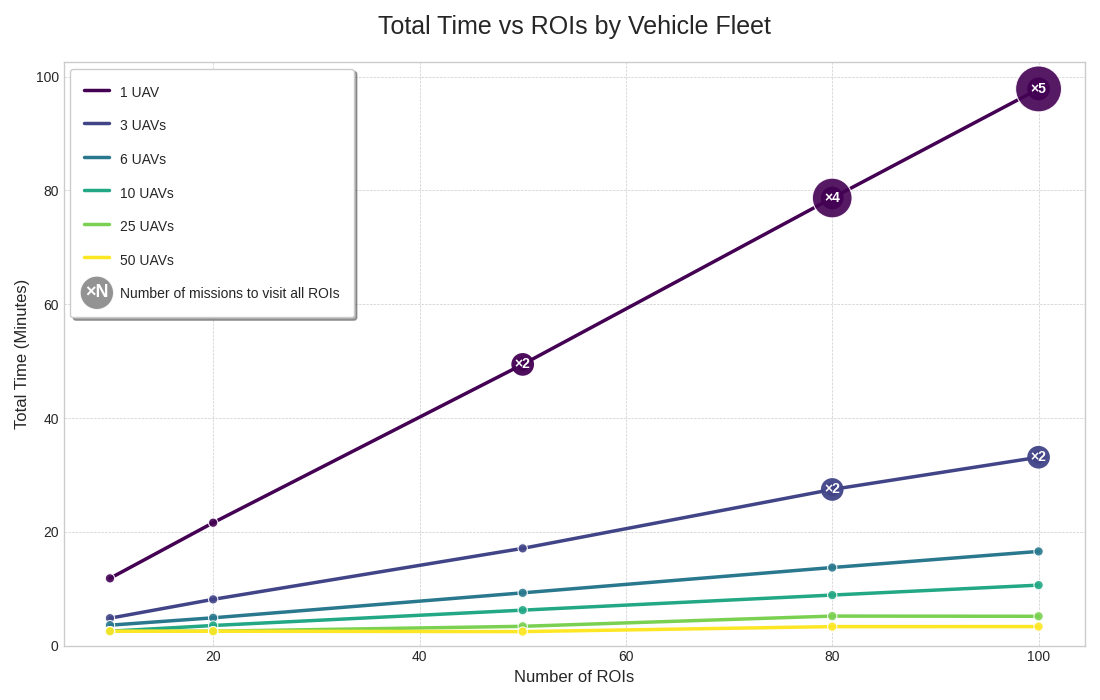}
	\caption{Mission duration as a function of swarm size and number of ROIs. Each curve corresponds to a different swarm size (1, 3, 6, 10, 25, 50 UAVs), and the y-axis reports the mission time required to visit all ROIs. Markers are annotated with ``×N'' when the swarm requires more than one mission per UAV to complete coverage (i.e., multiple sequential missions are needed to visit all ROIs).}
	\label{fig:multi-uav study}
\end{figure*}
For the multi-UAV study, the scenarios were crafted in a similar fashion to \ref{subsubsec:View} by perturbing a nominal latitude/longitude center for each one of the random polygons by ±0.003° in both latitude and longitude, producing compact but realistic clusters of ROIs. Each UAV is modeled with a horizontal speed of 10 m/s, a vertical speed of 3 m/s, and approximately 25 minutes of usable battery per mission. A mission is defined as one continuous flight on one battery: the UAV takes off, visits its assigned ROIs, and lands. For each experiment, ROIs were distributed among various numbers of available UAVs (1, 3, 6, 10, 25, 50), and we measured how long the mission takes as the number of ROIs increases from 10 to 100. The results are depicted in Figure \ref{fig:multi-uav study}. The plotted mission time corresponds to one deployment window. If the swarm cannot visit all ROIs within a single mission per UAV, the point is annotated with “×N,” where N is the number of missions that swarm needs in order to visit all ROIs.

The results show a clear trend with fleet size. With a single UAV, mission duration increases from roughly 12 minutes at 10 ROIs to about 100 minutes at 100 ROIs. In these higher-load cases, one UAV is not able to clear the full ROI set in a single mission, so it must fly multiple missions back-to-back\footnote{This calculation neglects the time required to swap batteries and assumes the operator has an ample supply of pre-charged batteries. In reality, both the swap and the UAV power cycle take non-zero time. The results should therefore be interpreted as a lower bound on the improvement achievable by adding more UAVs under these idealized conditions.}; this is indicated on the plot by labels such as ×2, ×4, and ×5. With three UAVs, the total mission time for 100 ROIs drops to approximately 33 minutes, which is a major reduction compared to the single-UAV case. However, for large ROI counts even three UAVs may still require more than one mission to complete inspection, and this appears as a ×2 label at the high end. Notably, from six UAVs onward the allocation/routing algorithm enable “one-shot” inspection: the entire ROI set is cleared within a single synchronized mission ($\sim$18 min for 100 ROIs), eliminating relaunches. This parallel servicing of many ROIs in one mission window achieves the ultimate reduction, a minimal field time footprint, directly translating to lower operational cost and exposure.

Beyond roughly 25 UAVs, the curves for 25 and 50 UAVs are nearly identical. This indicates diminishing returns: once each UAV is responsible for only a very small number of ROIs in a compact operating area, adding even more UAVs does not significantly shorten total mission time, because the dominant cost becomes basic transit time, not lack of parallelism. This diminishing-return behavior is scenario dependent. It is a consequence of the ROIs being relatively clustered in this experiment. If the ROIs were allowed to spread farther apart (i.e., if we increased the perturbation limits), then adding more UAVs would keep providing benefits for longer, because multiple vehicles could service distant clusters in parallel. In the limiting case where ROIs are extremely far apart, a UAV might only be able to reach and inspect a single ROI per mission. In that regime, the break-even point shifts: we would continue to gain by adding UAVs until you are effectively at one UAV per ROI.

Table \ref{tab:uav_dist_batt} summarizes, for each swarm size and ROI count, the average per-UAV flight distance (D) and battery usage (B) required to visit all the assigned ROIs within a single mission. The “--” indicates that the swarm could not visit all ROIs in a single mission and would require multiple missions.

The results show a clear scaling effect: larger swarms dramatically reduce the per-UAV mission load. At 20 ROIs, a single UAV must on average fly over 12 km and use 86\% of its battery, while a 3-UAV swarm brings this down to 4.5 km and 33\%, and a 6-UAV swarm brings it further down to 2.7 km and 20\%. This pattern continues at higher task densities. In the 100-ROI scenario, a 6-UAV swarm requires on average 9.2 km and 66\% battery per UAV in a single mission, whereas swarms of 25 or 50 UAVs reduce that to about 2.9 km/$\sim$21\% and ~1.9 km/$\sim$13\%, respectively. These larger swarms are also the first that can complete scenarios with 80 or 100 ROIs in one mission per UAV (i.e., without needing to relaunch). The difference between 25 and 50 UAVs is relatively small, suggesting diminishing returns as previously noted.

\begin{table*}[!ht]
\centering
\captionsetup{justification=centering}
\caption{Average \textbf{D}istance (m) and \textbf{B}attery consumption (\%) across UAV swarm sizes and numbers of ROIs. ``--'' indicates that more than one mission was required to visit all ROIs.}
\label{tab:uav_dist_batt}

%\setlength{\tabcolsep}{4pt}    % tighten column spacing
%\footnotesize                 % slightly smaller font

\begin{tabular}{@{}c*{10}{r}@{}}
\toprule
\multicolumn{1}{c}{\textbf{\# UAVs}} &
\multicolumn{10}{c}{\textbf{Number of ROIs}} \\
\cmidrule(lr){2-11}
& \multicolumn{2}{c}{\textbf{10}} &
  \multicolumn{2}{c}{\textbf{20}} &
  \multicolumn{2}{c}{\textbf{50}} &
  \multicolumn{2}{c}{\textbf{80}} &
  \multicolumn{2}{c}{\textbf{100}} \\
& \multicolumn{1}{c}{\textit{D(m)}} & \multicolumn{1}{c}{\textit{B(\%)}} 
& \multicolumn{1}{c}{\textit{D(m)}} & \multicolumn{1}{c}{\textit{B(\%)}} 
& \multicolumn{1}{c}{\textit{D(m)}} & \multicolumn{1}{c}{\textit{B(\%)}} 
& \multicolumn{1}{c}{\textit{D(m)}} & \multicolumn{1}{c}{\textit{B(\%)}} 
& \multicolumn{1}{c}{\textit{D(m)}} & \multicolumn{1}{c}{\textit{B(\%)}} \\
\midrule
1  & 6596 & 47.28 & 12052 & 86.39 &   --  &   --   &   --  &   --   &   --  &   --   \\
3  & 2687 & 19.26 &  4535 & 32.51 &  9536 & 68.36  &   --  &   --   &   --  &   --   \\
6  & 2003 & 14.36 &  2725 & 19.53 &  5173 & 37.08  &  7657 & 54.89  &  9237 & 66.22  \\
10 & 1409 & 10.10 &  1970 & 14.12 &  3476 & 24.91  &  4955 & 35.52  &  5929 & 42.50  \\
25 & 1409 & 10.10 &  1408 & 10.10 &  1891 & 13.55  &  2894 & 20.75  &  2874 & 20.60  \\
50 & 1409 & 10.10 &  1408 & 10.10 &  1408 &  10.10  &  1861 & 13.34  &  1863 & 13.36  \\
\bottomrule
\end{tabular}
\end{table*}

\section{Real-world Deployment}
\label{sec:real}
Section \ref{sec:sim} assumes ideal conditions that are rarely encountered in real-world operations. In this section, mUDAI is deployed in two real-world missions to assess its efficiency under realistic environments. One of the main differences between the simulations and real deployments concerns the terrain complexity. In Section \ref{sec:sim}, all experiments are conducted assuming flat terrain with no elevation variation, meaning that the UAV’s altitude and its distance from the ground, calculated during viewpoint optimization, are identical. However, such conditions are extremely rear to meet in actual testbeds. To address this issue, the flight altitude at each viewpoint is adjusted according to the local ground elevation, using data retrieved from precise elevation APIs\footnote{\url{https://developers.google.com/maps/documentation/elevation/overview}}. This section therefore aims to evaluate the applicability of the proposed method to real-world operations, and to identify unexpected deviations resulting from realistic environmental conditions. Moreover, a comparison with a multi-UAV Coverage Path Planning (mCPP) method is included, serving as a benchmark against what the authors consider the most relevant alternative approach for the specific problem.
\color{black}

\subsection{Experimental Setup}
\label{subsec:setup}
\begin{figure}[!ht]
    \centering
	\includegraphics[width=.9\linewidth]{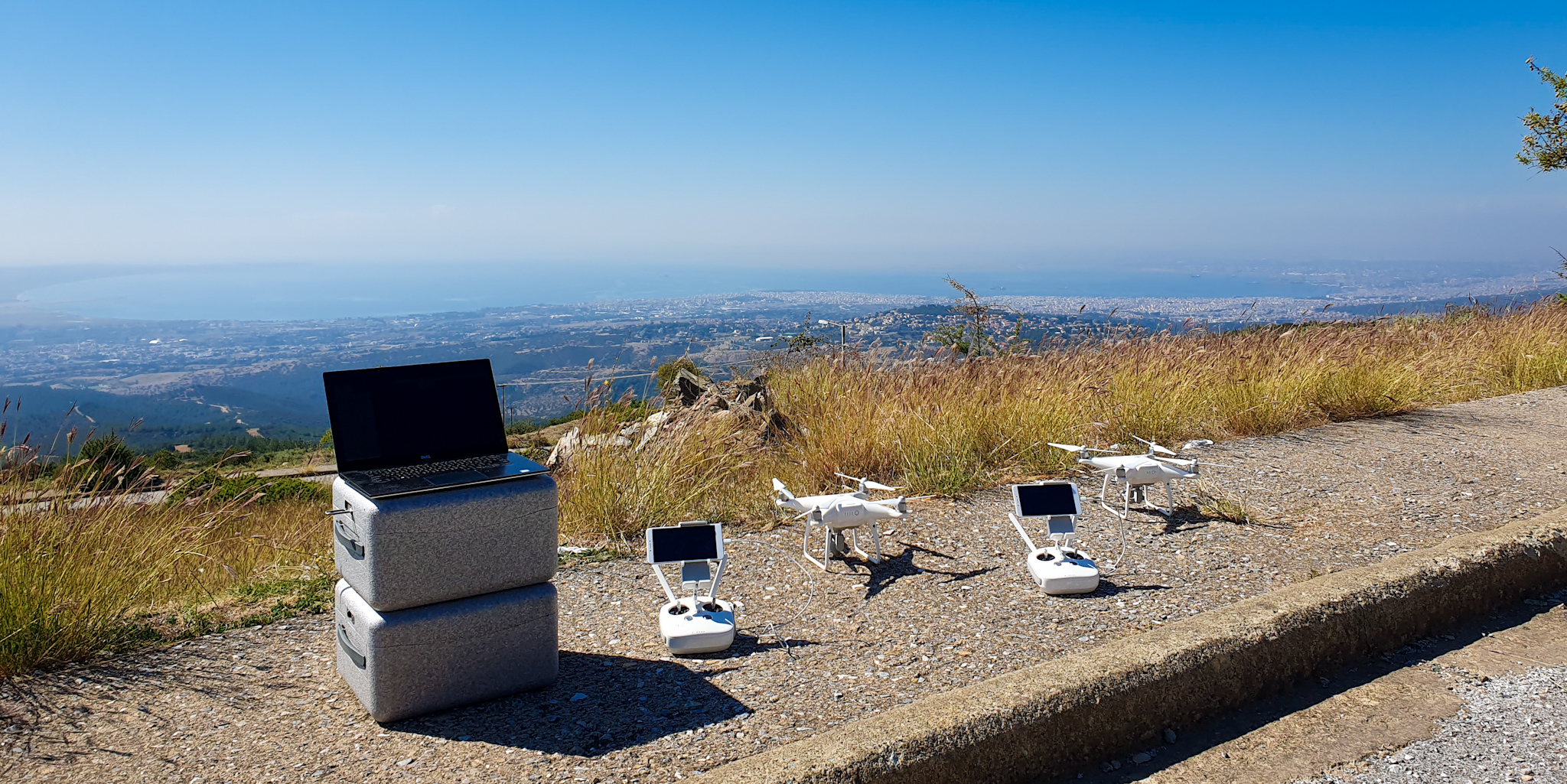}
	\caption{Experimental hardware setup used to validate the performance of the mUDAI algorithm. The setup includes a standard laptop and two commercial UAVs, each equipped with controllers running custom Android applications designed for mission execution and data collection.}
	\label{fig:setup}
\end{figure}

For the deployment of the mUDAI methodology in real-world scenarios, a setup similar to the one described in \cite{apostolidis2022cooperative} was used. Specifically, a set of commercial UAVs\footnote{\url{https://www.dji.com/phantom-4-pro}}, along with a custom Android application developed for this purpose, were used to execute all missions presented in sub-sections \ref{subsec:exp-Galatsi} and \ref{subsec:exp-Kissos}. Figure \ref{fig:setup} shows the exact hardware setup used for the deployment of the missions.

\subsection{Small-Scale Mission in Galatsi, Athens}
\label{subsec:exp-Galatsi}

This scenario consists of two separate mission deployments, in the same area, for the exact same five (5) ROIs to be inspected, using the two different optimization objectives for the calculation of the optimal viewpoint (MCO, BCO) in each of them. The overall ground elevation variations within this testbed are less than 10 meters, allowing for a safe and reliable validation of the flight altitude adjustment pipeline described above. \color{black} For the deployment of these missions a single UAV was used.

\begin{figure*}[!t]
  \centering
  \begin{tabular}{ p{0.325\textwidth} @{\hspace{0.01\textwidth}} p{0.325\textwidth} @{\hspace{0.01\textwidth}} p{0.325\textwidth}}
    \begin{minipage}{\linewidth}
      \centering
      \includegraphics[width=\linewidth]{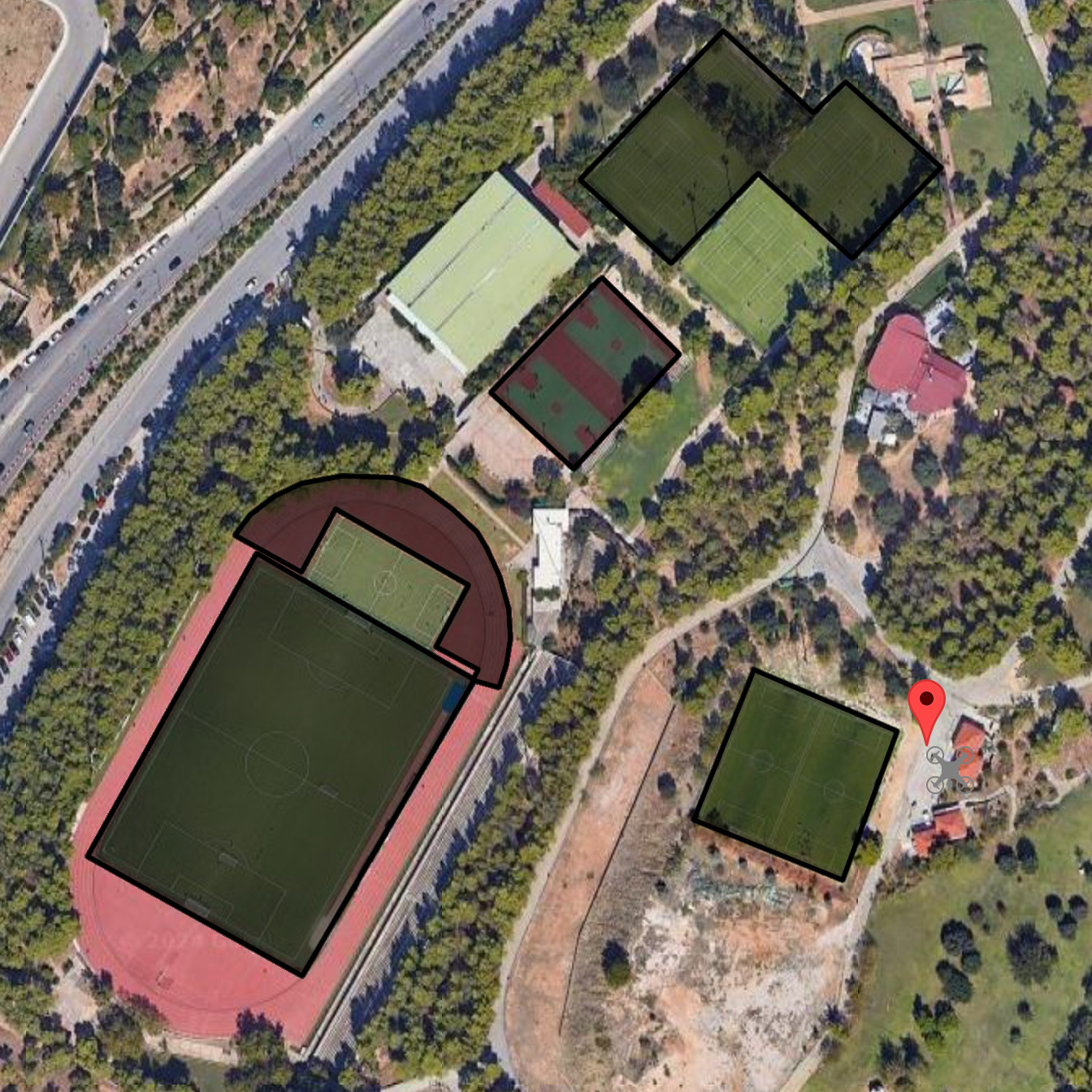} \\
      \small (a) FISR problem \\
    \end{minipage} &

    \begin{minipage}{\linewidth}
      \centering
      \includegraphics[width=\linewidth]{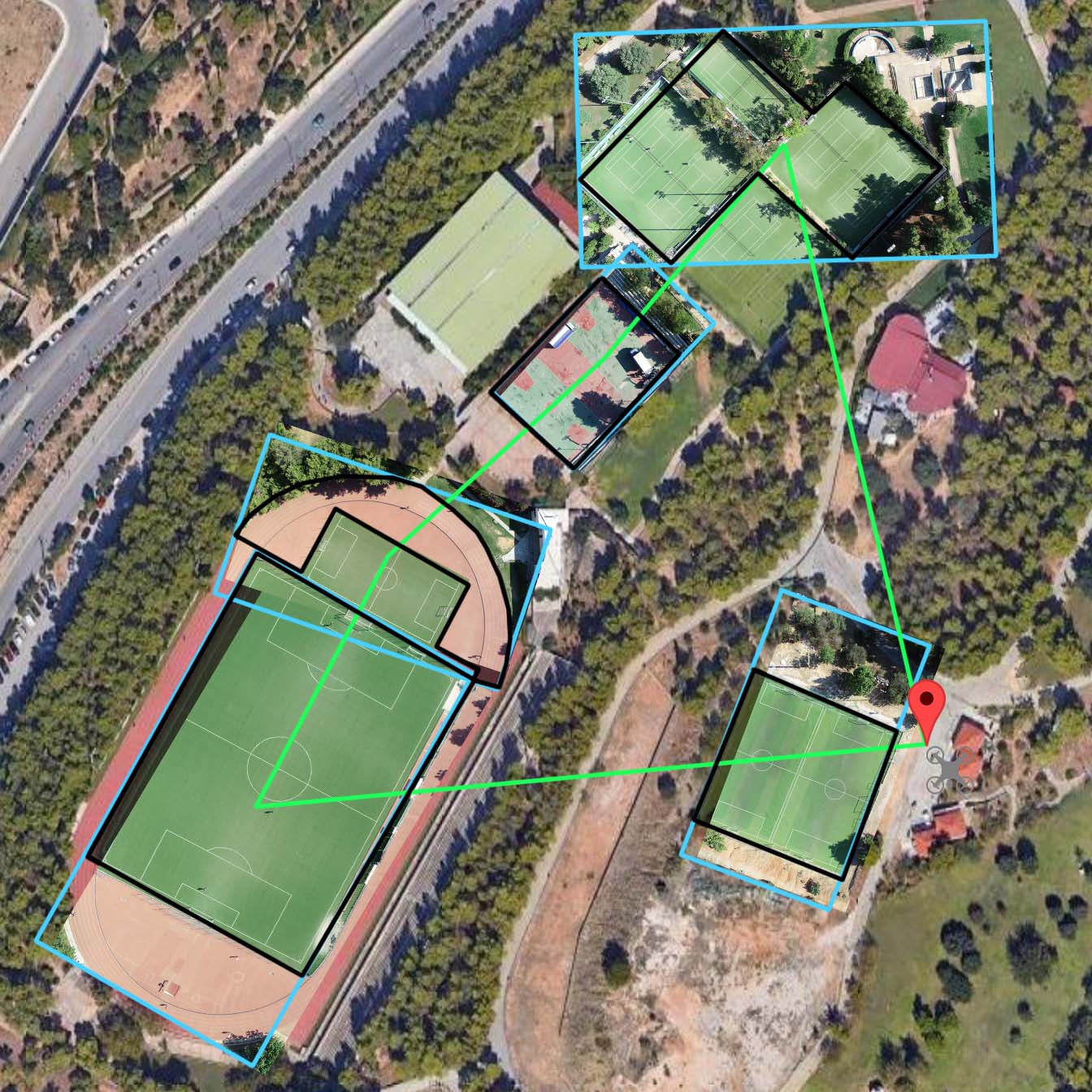} \\
      \small (b) MCO solution \\
    \end{minipage} &

    \begin{minipage}{\linewidth}
      \centering
      \includegraphics[width=\linewidth]{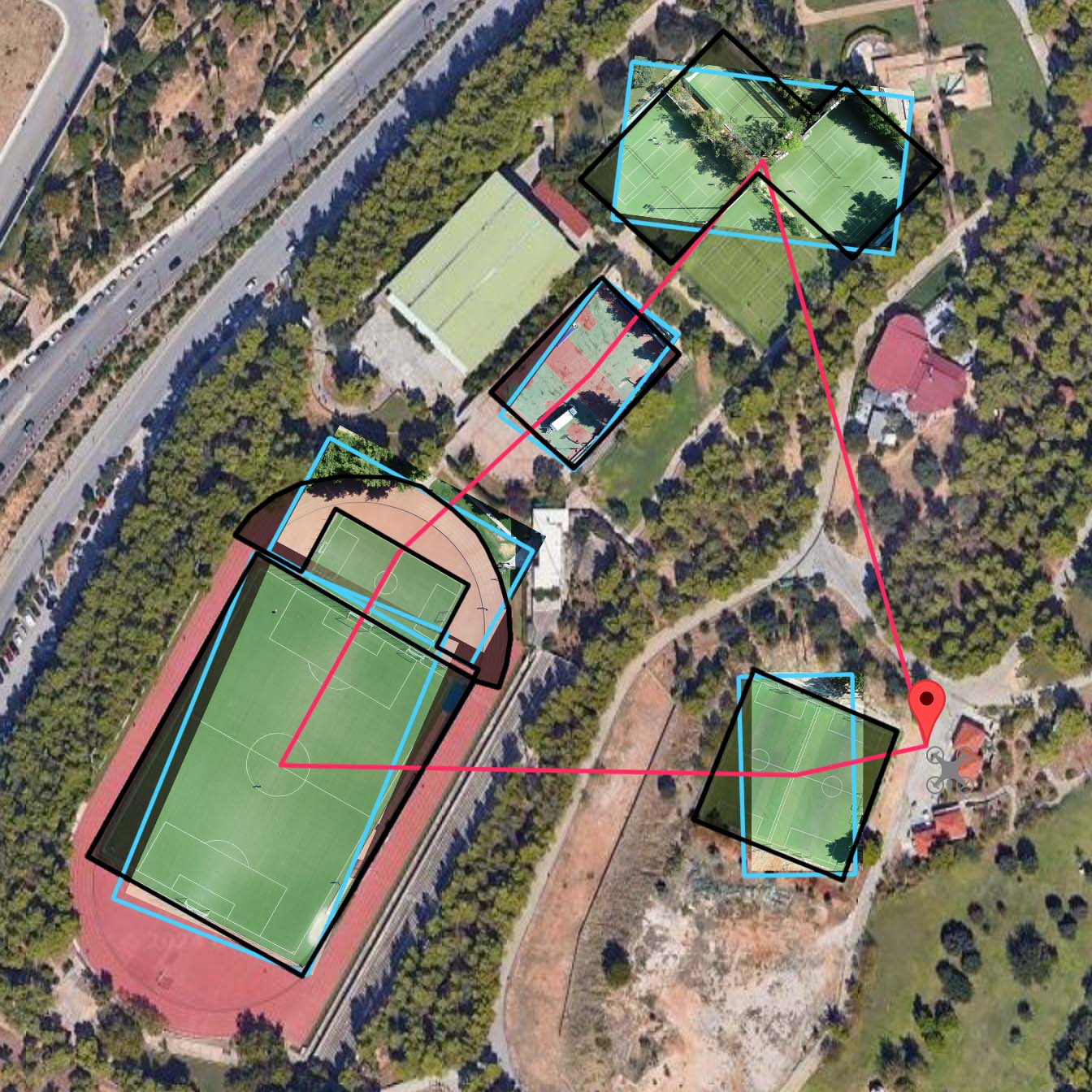} \\
      \small (c) BCO solution \\
    \end{minipage}
  \end{tabular}

  \medskip

  \caption{Real-world deployment of mUDAI solutions on testbed \#1, shown on a base map that provides geographic context for the area, with overlaid collected data. (a) FISR: The initial problem formulation with user-defined ROIs (black polygons) and the UAV's deployment position (red pin). (b) MCO: The optimized solution that ensures image captures containing the whole ROIs, where the calculated camera captures are illustrated with blue polygons, and the collected data are overlaid on the map. (c) BCO: The optimized solution that provides lower GSD in the collected images, but does not capture the whole ROIs, with the camera captures similarly illustrated by blue polygons, and the collected data overlaid on the map. In both (b) and (c), the trajectories, shown as connecting lines, represent the paths calculated by the VRP solver.}
  \label{fig:results-glac2}
\end{figure*}

As shown in figure \ref{fig:results-glac2}-(a), the user defines the FISR problem to be solved by selecting the ROIs and the initial position for the deployment of the UAV. Additionally, they define the minimum and the maximum flight altitude allowed for the viewpoints, and the designated flight altitude for the transition among these positions, the optimization objective that should be used for the calculation of the viewpoints, the number of UAVs that they intend to use, and the desired UAV speed, needed for the time estimation of the missions.

The main objectives of the mission deployments in this scenario is to visually explain the difference between MCO and BCO, to discuss the errors and deviations introduced in the collected data, and to provide a validation for the time estimation of the missions, a parameter critical for their safe deployment, since each mission should always respect the nominal operational duration of the UAVs used.

As shown in figure \ref{fig:results-glac2}-(b) and \ref{fig:results-glac2}-(c), MCO objective indeed calculates viewpoints that contain the whole ROIs, while simultaneously minimizing the area of non-designated regions in the camera capture, and BCO objective calculates viewpoints that allow for an image capture containing a large amount of - but not the whole - ROIs, managing this way to exclude larger amount of non-designated areas from the camera capture. From the overlaid collected data over the map, it is clear that there is a very small deviation among the calculated camera captures and the collected data, indicating that the flight altitude adjustment--according to the terrain elevation--works as expected. The observed deviation is mostly caused by four (4) factors: i) the drone's camera gimbal not facing vertically to the ground surface ($87\degree$ is the max allowed angle for the equipment used), ii) errors in the map that the regions were initially defined, iii) GPS errors, iv) deviations caused by wind gusts during the captures. 

Finally, we also assess the deviation between estimated and real time duration of the missions. The estimated time is calculated using the formula: 

\begin{equation}
    EstimatedDuration = \frac{Length}{Speed} + Turns \cdot \frac{c_1 \cdot Speed}{c_2 + |Speed|}
\end{equation}

The first component represents the ideal time required to complete the mission if the UAV maintained a constant speed throughout. The second component introduces an additional delay for each turn in the flight path. The constants $c_1$ and $c_2$ are parameters that can be adjusted to fine-tune the sigmoid function, ensuring it reflects the specific behavior and characteristics of the UAV. In this study, $c_1$ and $c_2$ are defined as 5 and 20, respectively. As shown in Table \ref{tab:glac2-time}, for the MCO mission deployment the real-world duration was 5.56\% smaller, and for the BCO 4.35\% smaller than the estimated ones. It is worth mentioning that having a slightly larger estimated than real-world mission duration is more desired than the opposite, since it provides a larger time margin for the safe return of the UAV to the base position, after the completion of the data collection. 

\begin{table}[!ht]   
    \centering
    \begin{tabular}{c|ccc}
         Time & Estimated & Real-world & Deviation\\
         \hline\hline
         MCO & 4.14 min & 3.91 min & 5.56\%\\
         \hline
         BCO & 4.14 min & 3.96 min & 4.35\%\\
    \end{tabular}
    \caption{Experimental results in testbed \#1}
    \label{tab:glac2-time}
\end{table}

\subsection{Large-Scale Mission in ``The Ghost City of Chortiatis'', Thessaloniki}
\label{subsec:exp-Kissos}

\begin{figure}[!t]
    \centering
	\includegraphics[width=.9\linewidth]{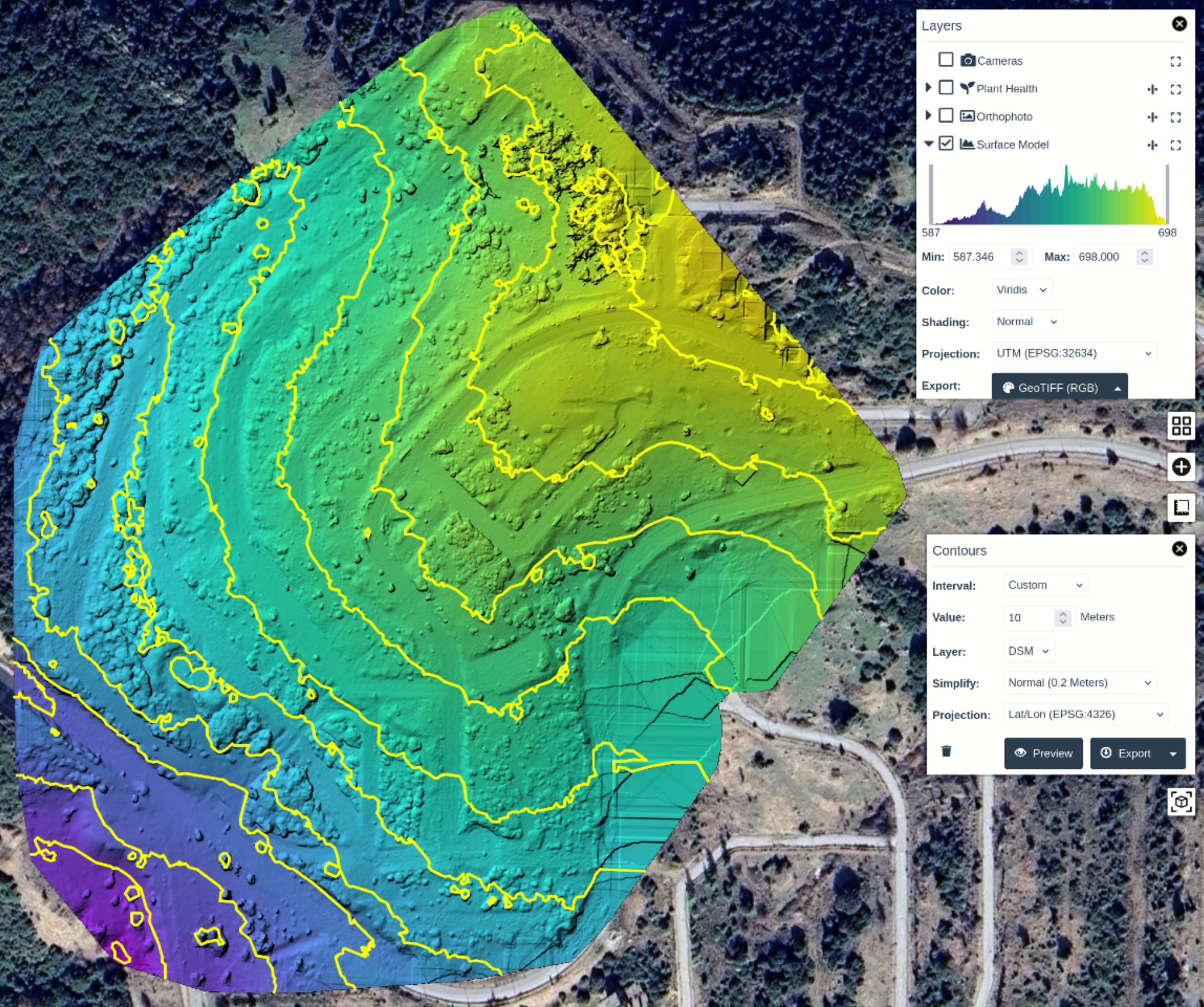}
	\caption{Surface model generated by the collected data for testbed \#2, indicating elevation variations over 110 meters within the operational area.\color{black}}
	\label{fig:elevation}
\end{figure}

\begin{figure*}[!t]
    \centering
    \begin{tabular}{p{0.42\linewidth} @{\hspace{0.04\linewidth}} p{0.42\linewidth}}
    \begin{minipage}{\linewidth}
      \centering
      \includegraphics[width=\linewidth]{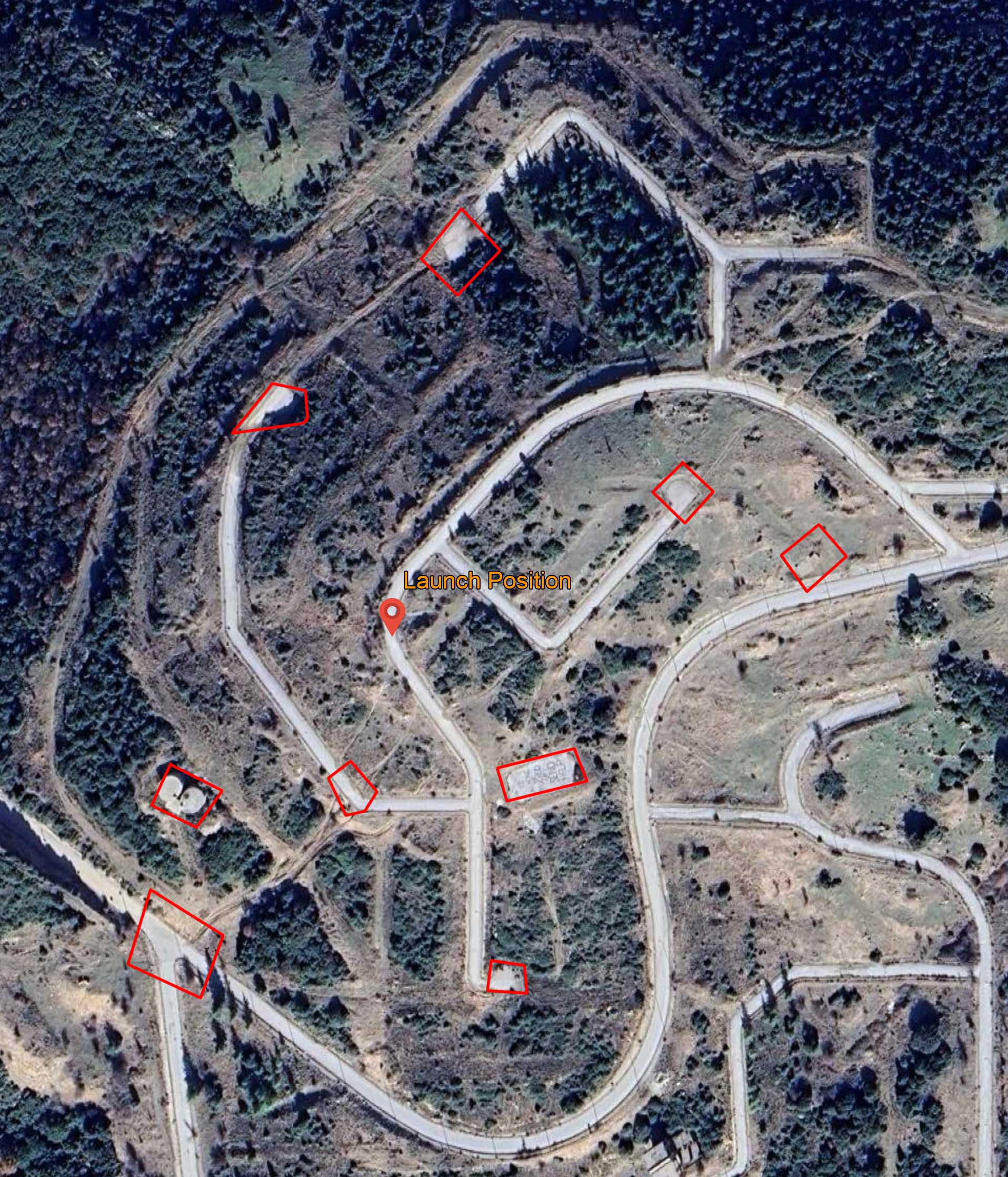} \\
      \small (a) FISR problem \\
    \end{minipage}
    
    \begin{minipage}{\linewidth}
      \centering
      \includegraphics[width=\linewidth]{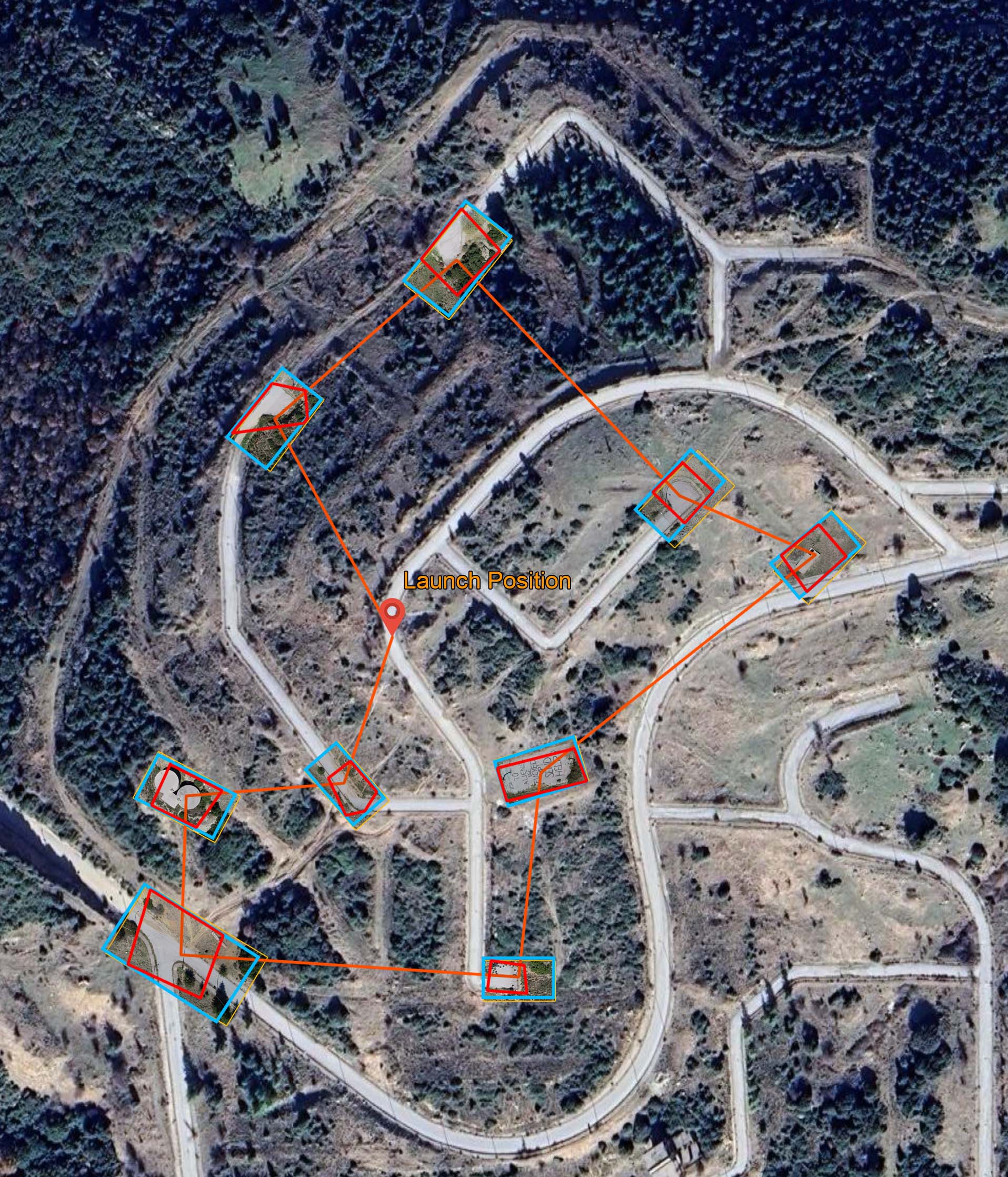} \\
      \small (c) mUDAI - 1 UAV \\
    \end{minipage} &
    
    \begin{minipage}{\linewidth}
      \centering
      \includegraphics[width=\linewidth]{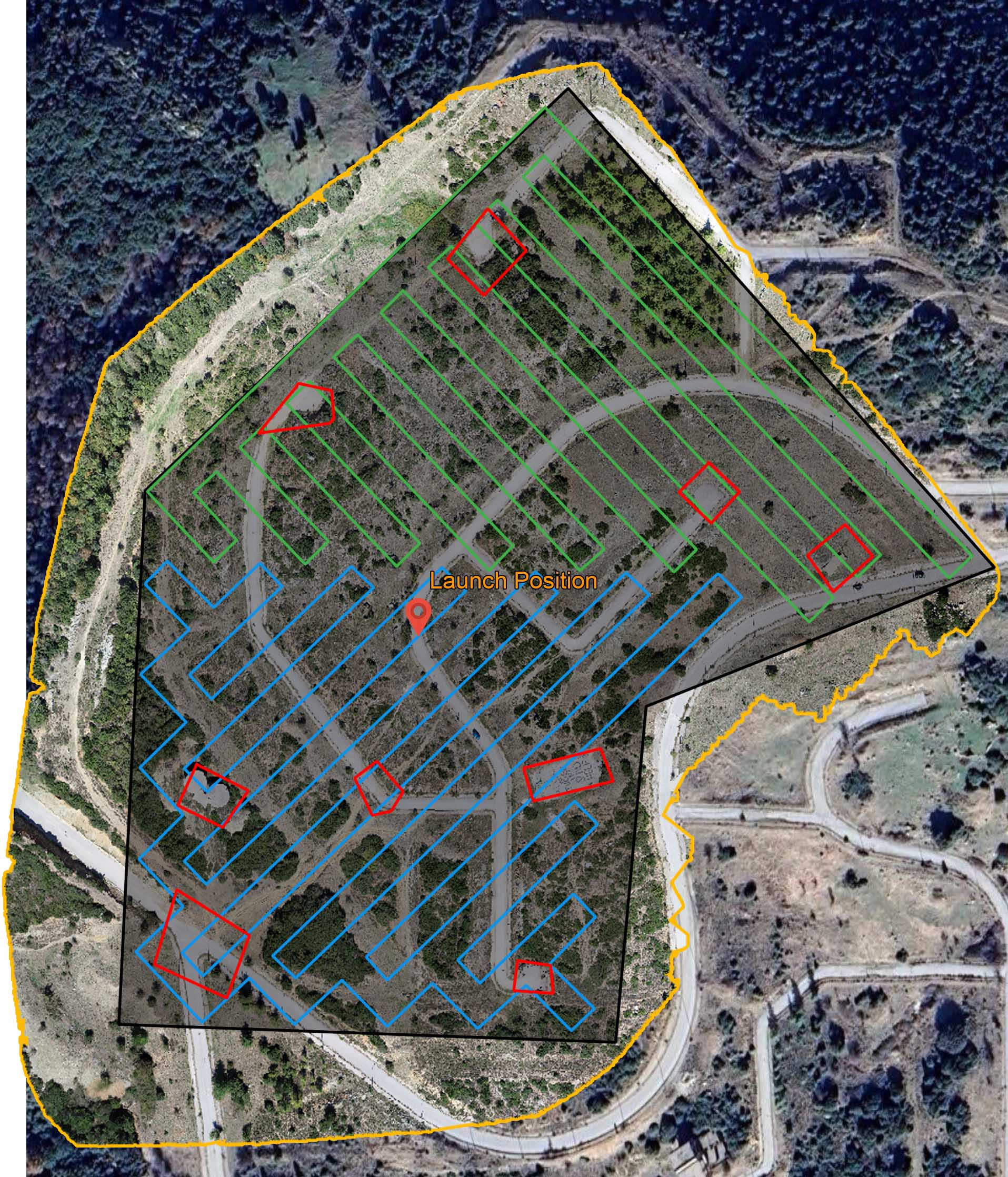} \\
      \small (b) mCPP \\
    \end{minipage}
    
    \begin{minipage}{\linewidth}
      \centering
      \includegraphics[width=\linewidth]{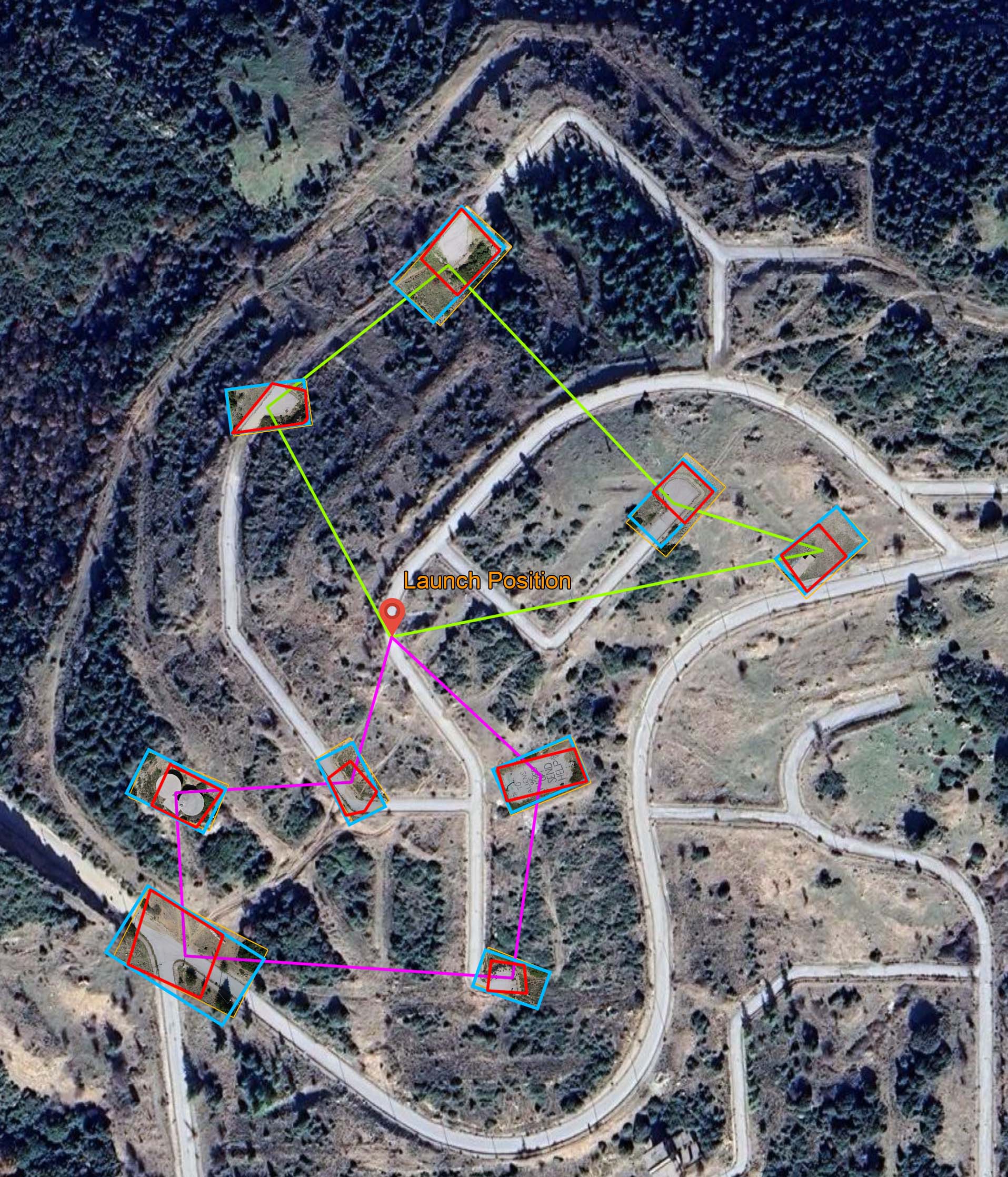} \\
      \small (d) mUDAI - 2 UAVs \\
    \end{minipage}
    \end{tabular}
    
    \medskip
    \caption{Real-world deployment of mCPP and mUDAI on testbed \#2. A map provides geographic context; data/orthomosaic are overlaid (yellow outline). (a) FISR: initial setup with user-defined ROIs (red polygons) and UAV start position (red pin). (b) mCPP: ROI (black polygon); collected data processed into orthomosaic (yellow outline). (c) mUDAI – 1 UAV: optimized single-UAV plan; blue polygons are camera footprints; collected data overlaid (yellow outline). (d) mUDAI – 2 UAVs: multi-UAV plan with distributed trajectories, camera footprints (blue), and collected data (yellow). In (b–d), colored lines denote trajectories from CPP algorithm and the VRP solver.}
    %\caption{Real-world deployment of mCPP and mUDAI solutions on testbed \#2, shown on a base map providing geographic context for the area, with overlaid collected data. (a) FISR: The initial problem formulation with user-defined ROIs (red polygons) and UAV deployment position (red pin). (b) mCPP: The ROI outlined by a black polygon, with the collected data processed to generate an orthomosaic - displayed on the map with a yellowish outline. (c) mUDAI - 1 UAV: The optimized solution for a single UAV, where the blue polygons represent the calculated camera captures, and the collected data are overlaid on the map - with a yellowish outline. (d) mUDAI - 2 UAVs: A multi-UAV solution, illustrating distributed trajectories, camera captures (blue polygons), and the collected data (with a yellowish outline). In (b), (c), and (d), the colorful lines represent the trajectories calculated by the CPP algorithm and the VRP solver, respectively.}
    \label{fig:results-ZEP-Kissos}
    
\end{figure*}

This scenario compares the use of a typical mCPP method\cite{apostolidis2022cooperative, chleboun2022improved}, deploying two UAVs to collect data out of the desired ROIs, with the use of mUDAI method for the same objective, using one and two UAVs in separate deployments. In this deployment the testbed was deliberately selected to have high terrain complexity and large elevation variations, to further stress the flight altitude adjustment procedure. As shown in Figure \ref{fig:elevation}, a surface model generated by the data collected during the mission, that elevation differences within the operational area exceed 110 meters. Figure \ref{fig:results-ZEP-Kissos}-(a) shows the FISR problem, as defined by the user over a map. In order to collect data out of these ROIs using a CPP method, with a flight altitude of 40m, a percentage of sidelap among sequential images of 75\% to successfully compose an orthomosaic of the covered region, and a UAV speed of 3m/s to allow for collection of non-corrupted images, the estimated mission duration for a single UAV is 35.5 minutes. This value exceeds the nominal operational duration of the UAV used, which is 25 minutes, making the deployment of two UAVs for the CPP mission mandatory. Regarding the mUDAI methodology, for a more detailed comparison the mission is deployed two times, one with a single UAV, since the mission's duration is smaller than the nominal operational duration of the UAV, and one with two UAVs, as in the CPP run. For both mUDAI deployments, the MCO optimization objective was used.

This experiment compares missions' duration and covered areas across all deployments, while also reporting the average GSD in each mission. For the qualitative assessment of coverage, we again employ the Recall and Precision metrics, as explained above. Table \ref{tab:ZEP-Kissos} presents all logged and calculated metrics, for each mission deployment. Additionally, figure \ref{fig:results-ZEP-Kissos} presents the defined FISR problem (\ref{fig:results-ZEP-Kissos}-(a)), and the trajectories along with the overlaid collected data for all three deployments (\ref{fig:results-ZEP-Kissos}-(b) - \ref{fig:results-ZEP-Kissos}-(d)).

\begin{table}[!ht]   
    {\centering
    \resizebox{\linewidth}{!}{
    \begin{tabular}{c|ccccc}
          & Time & Coverage Recall & Coverage Precision & Average GSD\\
         \hline\hline
         mCPP & 18.5 min & 100\% & 5\% & 1.09 cm/px$^*$\\
         \hline
         mUDAI \\ 1 UAV & 10.0 min & 99.3\% & 59.8\%  & 0.68 cm/px\\
         \hline
         mUDAI\\ 2 UAVs & 5.5 min & 99.3\% & 59.8\%  & 0.68 cm/px\\
    \end{tabular}}
    \caption{Experimental results in testbed \#2}
    \label{tab:ZEP-Kissos}}
    $^*$ While this value should be true and constant for a CPP mission deployed on a flat ground surface, since the operational altitude of the mission is not adjusted according to the topology of the region, this is just a theoretical value and does not correspond to the real GSD of the collected data.
\end{table}

The results demonstrate that while both mCPP and mUDAI achieve nearly identical coverage recall (100\% for mCPP and 99.3\% for mUDAI - where this small deviation is caused by the real-world errors introduced during mission deployments, as mentioned in the previous sub-section), mUDAI significantly outperforms mCPP in avoiding the collection of superfluous data. mCPP has a very low coverage precision of just 5\%, meaning a large portion of the captured data includes non-designated areas. In contrast, mUDAI - with both 1 and 2 UAVs - improves precision dramatically to 59.8\%, as it minimizes unnecessary data collection by focusing more effectively on the desired areas. Moreover, while mCPP could potentially achieve a similar or even better GSD, doing so would require a significant increase in mission time. Finally, regarding the flight adjustment according to the elevation of the terrain, used for the mUDAI deployment, the minimal deviation between the calculated camera captures and the collected images indicate that--if the elevation information used for the adjustment is precise enough--this methodology is capable of providing accurate results, even in highly-complex terrains.

Reducing irrelevant data collection is crucial for efficient post-processing, saving time and computational resources (for the generation of the orthomosaic, more than 3.5 hours were needed). mUDAI, especially with multiple UAVs, not only completes missions faster but also improves data quality by minimizing unnecessary captures. This precision enhances resource management, leading to quicker analysis and lower energy consumption, aligning with mUDAI’s goal of optimizing UAVs'mission efficiency.

\section{Conclusion, Limitations, \& Future Plans}
\label{sec:Conclusions}
This work introduced the Fast Inspection of Scattered Regions (FISR) problem and proposed a novel solution: the multi-UAV Disjoint Areas Inspection (mUDAI) methodology. Through extensive simulated evaluations, and experiments in both small and large-scale real-world scenarios, we demonstrated the effectiveness of mUDAI in optimizing UAV flight paths for the rapid inspection of multiple scattered regions. The results from the experiments clearly demonstrate the advantages of mUDAI over standard CPP techniques, particularly in missions where time efficiency is critical. By minimizing the area of non-designated regions in the UAV’s camera captures and optimizing the UAVs' trajectories, mUDAI proved capable of completing missions faster, while also being more efficient in terms of resources management, such as energy consumption, and computational time/cost needed for results' processing.

The formulation presented in this work is subject to two inherent limitations. First, the single-image-per-ROI assumption, while efficient, restricts flexibility. In principle, increasing UAV altitude would allow full coverage of arbitrarily large ROIs, however, regulatory altitude restrictions constrain this option. Within these bounds, wide ROIs may exceed the sensor footprint, forcing a trade-off between coverage and ground sampling distance (GSD). Higher altitudes improve coverage but inevitably reduce spatial resolution. Second, the geometric mismatch between the rectangular camera footprint and irregular ROI shapes can lead to inefficiencies. For elongated or non-convex ROIs, the BCO may compress the ROI and leave parts uncaptured, while the MCO captures the full ROI but introduces large redundant out-of-ROI regions. This reflects an intrinsic limitation of approximating complex polygons with single rectangular projections. 

As for future plans, we aim to expand the scope of this research to a series of relevant applications, such as disaster response, fire evolution monitoring, critical infrastructure monitoring and precision agriculture. Another promising direction, that could also contribute to overcoming the existing limitations of the method, is relaxing the single-viewpoint constraint toward multi-view inspection, enabling richer coverage of complex or elongated ROIs.

\section*{Acknowledgment}	
This research was funded by European Union’s Horizon Europe Research and Innovation Programmes PERIVALLON, under Grant Agreement No 101073952, and iDriving, under Grant Agreement No 101147004. Views and opinions expressed are however those of the authors only and do not necessarily reflect those of the European Union or CINEA. Neither the European Union nor the granting authority can be held responsible for them. We also gratefully acknowledge the support of NVIDIA Corporation with the donation of GPUs used for this research.

\bibliographystyle{elsarticle-num} 
\bibliography{refs}

@article{shakhatreh2019unmanned,
  title={Unmanned aerial vehicles (UAVs): A survey on civil applications and key research challenges},
  author={Shakhatreh, Hazim and Sawalmeh, Ahmad H and Al-Fuqaha, Ala and Dou, Zuochao and Almaita, Eyad and Khalil, Issa and Othman, Noor Shamsiah and Khreishah, Abdallah and Guizani, Mohsen},
  journal={Ieee Access},
  volume={7},
  pages={48572--48634},
  year={2019},
  publisher={IEEE}
}

@article{galceran2013survey,
  title={A survey on coverage path planning for robotics},
  author={Galceran, Enric and Carreras, Marc},
  journal={Robotics and Autonomous systems},
  volume={61},
  number={12},
  pages={1258--1276},
  year={2013},
  publisher={Elsevier}
}

@article{flood1956traveling,
  title={The traveling-salesman problem},
  author={Flood, Merrill M},
  journal={Operations research},
  volume={4},
  number={1},
  pages={61--75},
  year={1956},
  publisher={INFORMS}
}

@book{toth2002vehicle,
  title={The vehicle routing problem},
  author={Toth, Paolo and Vigo, Daniele},
  year={2002},
  publisher={SIAM}
}

@article{almadhoun2019survey,
  title={A survey on multi-robot coverage path planning for model reconstruction and mapping},
  author={Almadhoun, Randa and Taha, Tarek and Seneviratne, Lakmal and Zweiri, Yahya},
  journal={SN Applied Sciences},
  volume={1},
  pages={1--24},
  year={2019},
  publisher={Springer}
}

@article{cabreira2019survey,
  title={Survey on coverage path planning with unmanned aerial vehicles},
  author={Cabreira, Tau{\~a} M and Brisolara, Lisane B and Paulo R, Ferreira Jr},
  journal={Drones},
  volume={3},
  number={1},
  pages={4},
  year={2019},
  publisher={MDPI}
}

@article{thibbotuwawa2020unmanned,
  title={Unmanned aerial vehicle routing problems: a literature review},
  author={Thibbotuwawa, Amila and Bocewicz, Grzegorz and Nielsen, Peter and Banaszak, Zbigniew},
  journal={Applied sciences},
  volume={10},
  number={13},
  pages={4504},
  year={2020},
  publisher={MDPI}
}

@article{apostolidis2022cooperative,
  title={Cooperative multi-UAV coverage mission planning platform for remote sensing applications},
  author={Apostolidis, Savvas D and Kapoutsis, Pavlos Ch and Kapoutsis, Athanasios Ch and Kosmatopoulos, Elias B},
  journal={Autonomous Robots},
  volume={46},
  number={2},
  pages={373--400},
  year={2022},
  publisher={Springer}
}

@article{apostolidis2023systematically,
  title={Systematically Improving the Efficiency of Grid-Based Coverage Path Planning Methodologies in Real-World UAVs’ Operations},
  author={Apostolidis, Savvas D and Vougiatzis, Georgios and Kapoutsis, Athanasios Ch and Chatzichristofis, Savvas A and Kosmatopoulos, Elias B},
  journal={Drones},
  volume={7},
  number={6},
  pages={399},
  year={2023},
  publisher={MDPI}
}

@article{gabriely2001spanning,
  title={Spanning-tree based coverage of continuous areas by a mobile robot},
  author={Gabriely, Yoav and Rimon, Elon},
  journal={Annals of mathematics and artificial intelligence},
  volume={31},
  pages={77--98},
  year={2001},
  publisher={Springer}
}

@article{luna2022fast,
  title={Fast multi-UAV path planning for optimal area coverage in aerial sensing applications},
  author={Luna, Marco Andr{\'e}s and Ale Isaac, Mohammad Sadeq and Ragab, Ahmed Refaat and Campoy, Pascual and Flores Pe{\~n}a, Pablo and Molina, Martin},
  journal={Sensors},
  volume={22},
  number={6},
  pages={2297},
  year={2022},
  publisher={MDPI}
}

@inproceedings{choset1998coverage,
  title={Coverage path planning: The boustrophedon cellular decomposition},
  author={Choset, Howie and Pignon, Philippe},
  booktitle={Field and service robotics},
  pages={203--209},
  year={1998},
  organization={Springer}
}

@article{ghaddar2020pps,
  title={PPS: Energy-Aware grid-based coverage path planning for UAVs using area partitioning in the presence of NFZs},
  author={Ghaddar, Alia and Merei, Ahmad and Natalizio, Enrico},
  journal={Sensors},
  volume={20},
  number={13},
  pages={3742},
  year={2020},
  publisher={MDPI}
}

@inproceedings{luna2023spiral,
  title={Spiral coverage path planning for Multi-UAV photovoltaic panel inspection applications},
  author={Luna, Marco Andr{\'e}s and Isaac, Mohammad Sadeq Ale and Fernandez-Cortizas, Miguel and Santos, Carlos and Ragab, Ahmed Refaat and Molina, Martin and Campoy, Pascual},
  booktitle={2023 International Conference on Unmanned Aircraft Systems (ICUAS)},
  pages={679--686},
  year={2023},
  organization={IEEE}
}

@article{xie2022multiregional,
  title={Multiregional coverage path planning for multiple energy constrained UAVs},
  author={Xie, Junfei and Chen, Jun},
  journal={IEEE Transactions on Intelligent Transportation Systems},
  volume={23},
  number={10},
  pages={17366--17381},
  year={2022},
  publisher={IEEE}
}

@article{xie2020path,
  title={Path planning for UAV to cover multiple separated convex polygonal regions},
  author={Xie, Junfei and Carrillo, Luis Rodolfo Garcia and Jin, Lei},
  journal={IEEE Access},
  volume={8},
  pages={51770--51785},
  year={2020},
  publisher={IEEE}
}

@article{alzahrani2020uav,
  title={UAV assistance paradigm: State-of-the-art in applications and challenges},
  author={Alzahrani, Bander and Oubbati, Omar Sami and Barnawi, Ahmed and Atiquzzaman, Mohammed and Alghazzawi, Daniyal},
  journal={Journal of Network and Computer Applications},
  volume={166},
  pages={102706},
  year={2020},
  publisher={Elsevier}
}

@article{krestenitis2024overcome,
  title={Overcome the Fear Of Missing Out: Active sensing UAV scanning for precision agriculture},
  author={Krestenitis, Marios and Raptis, Emmanuel K and Kapoutsis, Athanasios Ch and Ioannidis, Konstantinos and Kosmatopoulos, Elias B and Vrochidis, Stefanos},
  journal={Robotics and Autonomous Systems},
  volume={172},
  pages={104581},
  year={2024},
  publisher={Elsevier}
}

@article{stefanopoulou2024improving,
  title={Improving time and energy efficiency in multi-UAV coverage operations by optimizing the UAVs’ initial positions},
  author={Stefanopoulou, Aliki and Raptis, Emmanuel K and Apostolidis, Savvas D and Gkelios, Socratis and Kapoutsis, Athanasios Ch and Chatzichristofis, Savvas A and Vrochidis, Stefanos and Kosmatopoulos, Elias B},
  journal={International Journal of Intelligent Robotics and Applications},
  pages={1--19},
  year={2024},
  publisher={Springer}
}

@incollection{dunbar2022remote,
  title={Remote Sensing: Satellite and RPAS (Remotely Piloted Aircraft System)},
  author={Dunbar, Martha Bonnet and Caballero, Isabel and Rom{\'a}n, Alejandro and Navarro, Gabriel},
  booktitle={Marine Analytical Chemistry},
  pages={389--417},
  year={2022},
  publisher={Springer}
}

@article{santiago2020comon,
  title={The Comon Agricultural Policy (CAP) and agroforestry systems in the European Union},
  author={Santiago Freijanes, Jos{\'e} Javier},
  year={2020}
}

@book{santos2014handbook,
  title={Handbook of optical sensors},
  author={Santos, Jos{\'e} Lu{\'\i}s and Farahi, Faramarz},
  year={2014},
  publisher={Crc Press}
}

@article{chleboun2022improved,
  title={An Improved Spanning Tree-Based Algorithm for Coverage of Large Areas Using Multi-UAV Systems},
  author={Chleboun, Jan and Amorim, Thulio and Nascimento, Ana Maria and Nascimento, Tiago P},
  journal={Drones},
  volume={7},
  number={1},
  pages={9},
  year={2022},
  publisher={MDPI}
}

@article{gao2024hierarchical,
  title={A hierarchical multi-UAV cooperative framework for infrastructure inspection and reconstruction},
  author={Gao, Chuanxiang and Wang, Xinyi and Chen, Xi and Chen, Ben M},
  journal={Control Theory and Technology},
  volume={22},
  number={3},
  pages={394--405},
  year={2024},
  publisher={Springer}
}

@article{liu2025advancing,
  title={Advancing Multi-UAV Inspection Dispatch Based on Bilevel Optimization and GA-NSGA-II},
  author={Liu, Yujing and Chen, Chunmei and Sun, Yu and Miao, Shaojie},
  journal={Applied Sciences},
  volume={15},
  number={7},
  pages={3673},
  year={2025},
  publisher={MDPI}
}
\end{document}